\pdfoutput=1
\documentclass[11pt]{article}

\usepackage{ACL2023}

\usepackage{times}
\usepackage{latexsym}
\usepackage{booktabs}

\usepackage[T1]{fontenc}
\usepackage[utf8]{inputenc}

\usepackage{microtype}

\usepackage{inconsolata}
\usepackage{multirow}

\usepackage{xcolor}
\definecolor{redish}{RGB}{223, 122, 094}
\definecolor{yellowish}{RGB}{242, 204, 142}
\definecolor{beige}{RGB}{244, 241, 222}
\definecolor{dkgreen}{RGB}{0, 150, 0}

\usepackage[]{listings}
\usepackage{CJKutf8}
\newcommand{\zh}[1]{\begin{CJK*}{UTF8}{gbsn}\small#1\end{CJK*}}
\usepackage[normalem]{ulem}
\usepackage{graphicx} 
\usepackage{enumitem}
\newcommand{\namedref}[2]{\hyperref[#2]{#1~\ref*{#2}}}
\newcommand{\sectionref}[1]{\namedref{Section}{#1}}
\newcommand{\tableref}[1]{\namedref{Table}{#1}}
\newcommand{\figureref}[1]{\namedref{Figure}{#1}}
\newcommand{\appendixref}[1]{\namedref{Appendix}{#1}}

\newcommand{\llama}{Llama2-7B}
\newcommand{\chatgpt}{GPT-3.5-turbo}
\newcommand{\gpt}{GPT-4}
\newcommand{\claude}{Claude-2}

\usepackage{gb4e}
\noautomath

\title{GEE! Grammar Error Explanation with Large Language Models}
\author{Yixiao Song$^{\spadesuit}$\thanks{~~~Work partially done as an intern at QuillBot.} \quad Kalpesh Krishna$^{\spadesuit}$\thanks{~~~Currently at Google.} \quad Rajesh Bhatt$^\spadesuit$ \\  {\bf Kevin Gimpel}$^{\heartsuit}$ \quad {\bf Mohit Iyyer}$^{\spadesuit}$ \vspace{8pt}\\
$^\spadesuit$University of Massachusetts Amherst\quad   $^\heartsuit$QuillBot \\ 
\texttt{\small \{yixiaosong,bhatt\}@umass.edu} \quad \texttt{\small miyyer@cs.umass.edu} \\ \texttt{\small  kevin.gimpel@quillbot.com} \quad \texttt{\small kalpeshk2011@gmail.com}
}

\begin{document}
\maketitle

\begin{abstract}

Grammatical error correction tools are effective at correcting grammatical errors in users' input sentences but do not provide users with \textit{natural language} explanations about their errors. Such explanations are essential for helping users learn the language by gaining a deeper understanding of its grammatical rules \citep{dekeyser2003implicit, ellis2006}.

To address this gap, we propose the task of \textit{grammar error explanation}, where a system needs to provide one-sentence explanations for each grammatical error in a pair of erroneous and corrected sentences. We analyze the capability of GPT-4 in grammar error explanation, and find that it only produces explanations for $60.2\%$ of the errors using one-shot prompting. 

To improve upon this performance, we develop a two-step pipeline that leverages fine-tuned and prompted large language models to perform structured atomic token edit extraction, followed by prompting \gpt\ to generate explanations. We evaluate our pipeline on German and Chinese grammar error correction data sampled from language learners with a wide range of proficiency levels. Human evaluation reveals that our pipeline produces $93.9\%$ and $98.0\%$ correct explanations for German and Chinese data, respectively. To encourage further research in this area, we will open-source our data and code.\footnote{\url{https://github.com/Yixiao-Song/GEE-with-LLMs}}

\end{abstract}

\section{Introduction}

Grammatical error correction (GEC) is a practical and valuable application of natural language processing that facilitates both proofreading of text and language learning. Recent advances in large language models (LLMs) have significantly improved the capabilities of GEC systems \citep{10.1145/3474840, bryant2022grammatical}; however, they are unable to \emph{explain errors in natural language} alongside providing  correction. 
% Error explanation is a functionality that is critical especially to language learners.
Error explanation is crucial to language learning and teaching \citep{ellis2010epilogue}: while corrections are a form of implicit feedback, they are not as impactful as explicit feedback, which involves pointing out errors and providing meta-linguistic information to the user (e.g., rules of forming well-formed phrases or sentences) \citep{dekeyser2003implicit, ellis2006}.  

% Current GEC models only shows learners how to correct an error but not why it is an error. From the perspective of second language acquisition, such corrections are only implicit feedback. While being useful, in the long term, it is not as impactful as explicit feedback, which involves pointing out the error(s) and providing meta-linguistic information (e.g., rules of forming well-formed phrases or sentences) \citep{dekeyser2003implicit, ellis2006}.\kkcomment{Move the previous 2-3 sentences to first paragraph, before defining GEE} The GEE task, built upon GEC tasks, aims to offer such explicit feedback. 

\begin{figure}[t]
    \centering
    \includegraphics[width=0.46\textwidth]{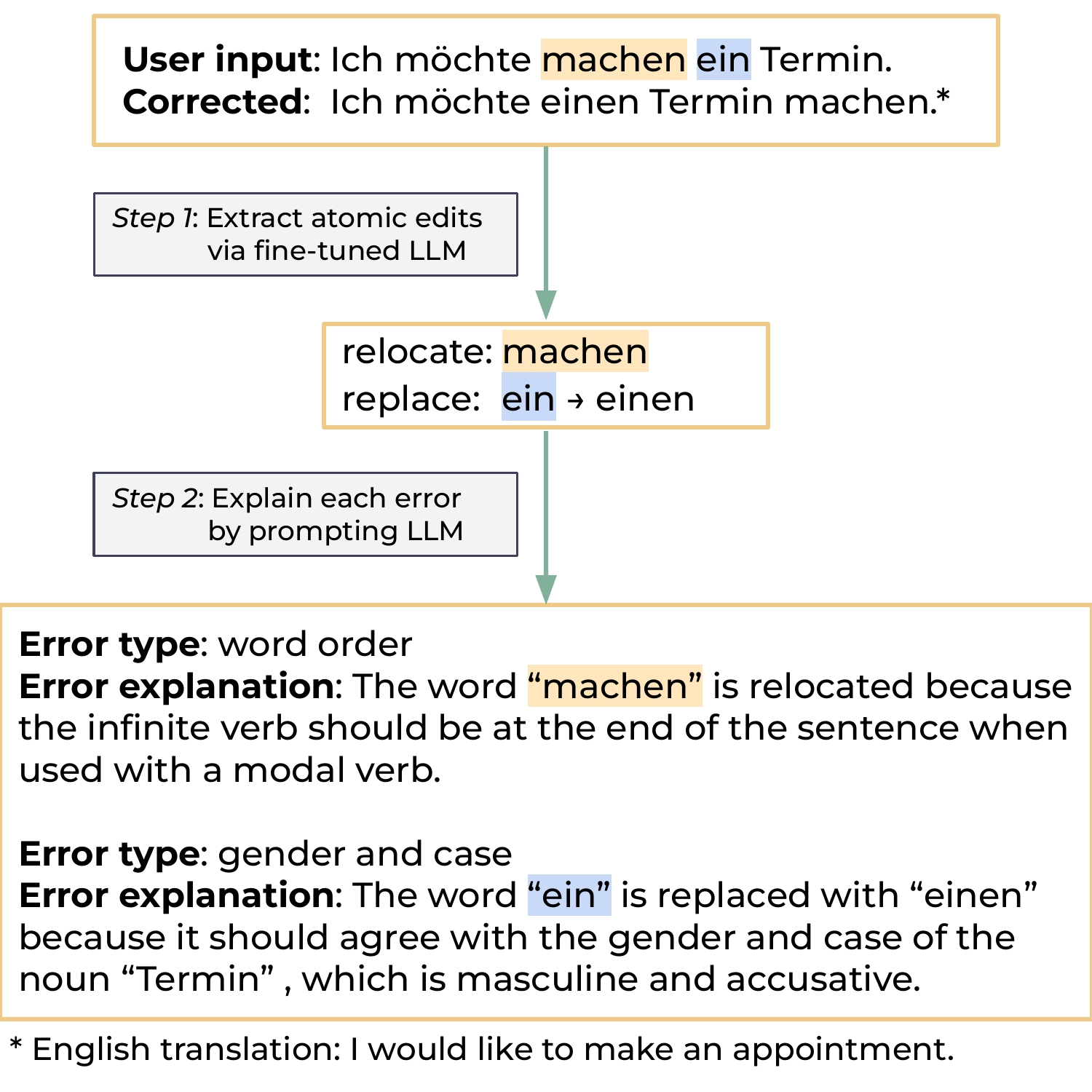}
    \caption{An illustration of the two-step pipeline of grammar error explanation (GEE). Given a pair of sentences with corrected errors, the GEE system first extracts linguistically meaningful edit units as errors. The extracted errors are then paired with the sentences as the input for GEE generation. Note: The error in \textit{einen} can be caused by \textit{gender} or \textit{case} or both. Without guessing the mental state of a language user, both are offered as the reasons in the explanation.}
    \label{fig:pipeline}
    \vspace{-0.15in}
\end{figure}
% https://docs.google.com/presentation/d/1pwvwk1theJU8jzJDeLWH0HhZcpOO3dsrka6Qzevpffc/edit?usp=sharing

In this work, we propose a new task---\emph{grammar error explanation} (GEE)---for which a model must generate natural language error explanations that help language learners acquire and enhance grammar knowledge. 
As shown in \figureref{fig:pipeline}, given a pair of sentences in which one sentence has grammar errors and the other one is corrected, a model needs to generate an explanation for each corrected grammar error. Given the capabilities of modern LLMs, can they simply solve this task via prompting? We show in~\sectionref{sec:gpt4_cant_end2end} that one-shot \gpt\ prompting detects only $60.2\%$ of the true errors and correctly explains only $67.5\%$ of the errors it does detect.

Given this result, we develop a pipeline for GEE generation that features an essential intermediate step---atomic token edit extraction. As shown in \figureref{fig:pipeline}, given an erroneous sentence and its corrected counterpart (source and target), we first extract atomic edits at the token level by prompting or fine-tuning LLMs such as \gpt, which also label the edits with one of four operation-level edit type labels: \texttt{insert}, \texttt{delete}, \texttt{replace}, and \texttt{relocation}. 
% \kgcomment{is this labeling done deterministically based on rules? or is it part of the output of the LLM doing the extraction?}
% \micomment{maybe use texttt for these edit types?} 
In the second step, we append the extracted edits to the source and target sentences and use them as the input to a GEE system. We utilize the few-shot learning ability of LLMs~\citep{brown2020language} to generate error explanations using carefully crafted language-specific prompts. 

% \micomment{this is too vague, how does this prompt interact with the atomic edits?} 

We validate our GEE pipeline on datasets in both German and Chinese, two languages with very different properties (fusional vs.\ analytical). We also recruit language teachers to evaluate the correctness of the explanations. 
For the first step in the pipeline, our atomic token edit method extracts $92.3\%$ of the true edits for German, which is $32.1\%$ higher than the naive one-shot approach in \sectionref{sec:gpt4_cant_end2end}. For the final GEE outputs in German, $93.9\%$ of the generated explanations are judged as correct by two German teachers. Similar performance is observed in Chinese with a $98\%$ correctness rate, suggesting that our two-step pipeline together with carefully crafted language-specific prompts generalizes well for different languages.

% For Chinese, two Chinese teachers judge $98\%$ of the outputs as correct. Despite the high correctness rate, through a careful analysis of the outputs and annotations, we found that \gpt~falls short in certain nuances in language grammar (e.g., \textit{-n} suffix for dative and plural in German).

% \yscomment{drop the chinese results from the intro and just replace with something like "we observe similar performance on chinese, suggesting our method generalizes across languages"}

% \micomment{don't just dump numbers like this, summarize the high level result (e.g., how much better is this than prompting GPT4?) and only include specific numbers if there is value in the reader knowing them right now} 

In summary, our contributions are the following. First, we propose a new task on grammar error explanation to enhance the utility of current grammatical error correction systems. Second, we propose a two-step pipeline and study its performance in German and Chinese with detailed error analysis. Third, we publicly release our atomic edit extraction datasets for German and Chinese as well as all LLM-generated GEE outputs with the aim of enabling future research on GEE and facilitating the development of more effective GEE systems.

\section{GEE task definition}\label{sec:task.def}

While most GEC models provide viable grammar error corrections \citep{bryant-ng-2015-far,bryant2022grammatical}, they do not provide explanations in natural language alongside the corrections, which are critical for language learners in mastering grammar \citep{ellis2006, ellis2010epilogue}. 
% \kgcomment{here and in the intro, i think we need to be careful about this claim. Bryant et al's work (ERRANT etc) can be described as providing explanations for grammar error corrections. we should state in Sec 1 or here in Sec 2 how what we're doing differs from ERRANT}
In this section, we propose and define the task of grammar error explanation, which aims to fill this gap. In this work, we assume that a GEE model has access to the outputs of an existing GEC model, which produces the corrected form of an ungrammatical input sentence.

\subsection{Formalizing the GEE task}

% \kkcomment{move error types from section 5.1 here IMO}

% \kkcomment{let $X_{error}$ be a sentence input by the user, which contains grammatical errors. Let $X_{correct} = GEC(X_{error})$ be the grammatically correct version of $X_{error}$ produced by an existing GEC system. We assume that $X_{error}$, $X_{correct}$ are provided as inputs to a GEE system.}

% \kkcomment{Let $c_1^X, c_2^X, ..., c_n^X$ be a list of corrections made by the grammar error correction system to $X_{error}$. We assume that there are four types of errors, and each $c_i^X$ is a tuple <add in error types stuff from Section 5.1>.}

% \kkcomment{Let $e_1^X$, $e_2^X$, ..., $e_n^X$ correspond to a single sentence natural language explanation for the different corrections $c_1^X, c_2^X, ..., c_n^X$. <give example mapping error type to explanation, ideally in a table>}

% \kkcomment{The overall goal of GEE is,\\
% \noindent\textbf{Input}: $X_{error}, X_{correct}$\\\textbf{Output}: $e_1^X$, $e_2^X$, ..., $e_n^X$.
% As a concrete example, consider \figureref{fig:pipeline} <add details on how the definition connects to Fig 1>
% }

The input to a GEE model is a pair of sentences\footnote{In principle, the inputs could also be documents, but we restrict our work to sentence-level GEE and leave longer texts to future work.} in which one has (potentially multiple) grammar errors and the other is corrected. Concretely, let $X_{\mathit{error}}$ be an input sentence written by an user which contains grammatical errors. Then, $X_{\mathit{correct}} = \mathit{GEC}(X_{\mathit{error}})$ is the grammatically correct version of $X_{\mathit{error}}$ produced by an existing GEC system. 
Following common practice in GEC research \citep{bryant-etal-2017-automatic,lee-etal-2018-building,rao-etal-2020-overview}, we assume that an error can be corrected in four ways: insert, delete, replace, and relocate. Let $c_1^X, c_2^X, ..., c_n^X$ be a list of corrections made by the GEC system to $X_{\mathit{error}}$ through one of these four types of edits. Then, the goal of GEE is to generate single-sentence explanations in natural language $s_1^X$, $s_2^X$, ..., $s_n^X$ corresponding to each of these edits $c_1^X, c_2^X, ..., c_n^X$ (example in \figureref{fig:pipeline}). Concretely,
\begin{center}
\textbf{Input}: $X_{\mathit{error}}, X_{\mathit{correct}}$\\\textbf{Output}: $s_1^X$, $s_2^X$, ..., $s_n^X$
\end{center}

% \subsection{What makes a good explanation?}
% \subsection{Finding Error Boundaries}
\subsection{Error extraction as foundation of GEE}
% \kgcomment{the title is about what makes for a good explanation, but then the text below is really about motivating atomic edit extraction and saying that in order to have good explanations, we need to ensure the atomic edits are extracted in a reasonable way. i think a broader title would help here.}

% \kgcomment{it's a little hard to understand whether this section is describing a problem, making claims (which require argument/evidence), or making recommendations (which require justification). i think part of this section is about the importance of atomic edit extraction and part of it is about specificity/applicability of explanations. if we want to get into atomic edits at this stage in the paper, i think we should give our definition of atomic edits. if we don't want to get into that yet, we should just say that for simplicity we are focusing on the situation where atomic edits have been extracted, deferring discussion of that for later. i personally would dive into atomic edits here because i think it's an important part of the problem and something that relates to specificity of explanations.} 
% \kgcomment{here's how i would say it: when an expert corrects an error and explains why the correction was made, they intuitively identify the ``atomic errors'' and explain each one from the perspective of the grammar and spelling conventions of the language.} 

The quality of error explanation depends on how the list of corrections $c_1^X, c_2^X, ..., c_n^X$ is defined. Consider the corrections in (\ref{ex.src.trg0}), one way to define the correction list is through a string-based transformation (i.e., replace \textit{machen ein termin} with \textit{einen Termin machen}). However, when being asked to explain such an error, one would naturally approach it as, for example, ``\textit{machen} is moved'', ``\textit{ein} is replaced with \textit{einen}'', and ``\textit{termin} is changed''. On the other hand, for the corrections made in (\ref{ex.src.trg1}), one would not explain the error word by word as ``\textit{with} is moved''.  This leads to our consideration of what defines a good correction list. 

\begin{exe}

    \ex\label{ex.src.trg0}

    \textbf{S}: Ich m\"ochte \colorbox{redish}{machen} \colorbox{yellowish}{ein} \colorbox{beige}{termin}.

    \textbf{T}: Ich m\"ochte \colorbox{yellowish}{einen} \colorbox{beige}{Termin} \colorbox{redish}{machen}.

    \ex\label{ex.src.trg1}

    \textbf{S}: I \colorbox{redish}{with my puppy} go to the store.

    \textbf{T}: I go to the store \colorbox{redish}{with my puppy}.

    % \textbf{E}: The most common position of an adverb of manner is after the verb phrase (e.g., \textit{go to the store}).
    
\end{exe}

When shown corrected texts, experts naturally tend towards explaining errors in an atomic way, which roughly amounts to one explanation per error. This also allows learners to follow and understand explanations better, especially when there are contiguous errors in the input. All of these require a process of atomic error extraction, such as the ones described for (\ref{ex.src.trg0}) and (\ref{ex.src.trg1}) which naturally uses the conventions of grammar, spelling, and language usage. In the GEE task, we define atomic errors based on how a human expert would explain a set of error corrections to a language user. In what follows, we treat each atomic error as an atomic edit and give a working definition of how to identify it.  

Using (\ref{ex.src.trg0}) as an example, an edit (\textit{machen ein termin}) should be divided into smaller edits (\textit{machen}, \textit{ein}, and \textit{termin}) if an expert would explain the edit as merely the concatenation of explanations for the smaller edits. These smaller edits are then atomic edits (i.e., each of which has its own distinct explanation). Similarly, if an expert would explain an edit with multiple words using one explanation that cannot be separated into the concatenation of an explanation per token, then that multi-word operation is an atomic edit. An example is the relocation of \emph{with my puppy} in (\ref{ex.src.trg1}).

The working definition of atomic edits is a guideline of extracting linguistically meaningful edits. However, language specific decisions might be needed for individual languages (e.g., verbs with a separable prefix in German). We discuss such details for German and Chinese in \sectionref{sec:pipeline} and \appendixref{appendix:manual_extract}.

% There can be cases where one token is split/merged into or replaced by two tokens or vice versa in the source and target sentences. In such cases, each error should include all split/merged/replaced tokens. Other specific decisions might be needed for individual languages (e.g., verbs with a separable prefix in German). More details of deciding error boundaries are discussed in \sectionref{sec:pipeline} and \appendixref{appendix:manual_extract}.

% \kkcomment{for someone skimming the paper it's not immediately obvious that the explanations below are bad. Can we make it more explicit, maybe replace (a) with Bad Explanation 1 and (b) with Bad Explanation 2}

% GEE outputs should also be informative. Generic explanations such as the ones in (\ref{ex.meaningless.expl}) are \textbf{less helpful} for a learner to know how to prevent the same error in the future because they do not point out the underlying meta-linguistic information.

% \begin{exe}
%     \ex\label{ex.meaningless.expl}
%     Generic explanations are \textbf{less helpful}:
%     \begin{xlist}
%         \ex\label{ex.meaningless.expl0}
%         The word X is wrong and the word Y is the correct word to use in this context.

%         \ex\label{ex.meaningless.expl1}
%         This is a preposition error.
%     \end{xlist}
% \end{exe}

% GEE can be further enhanced by providing examples with errors similar to the ones that a learner makes. However, this is beyond the scope of our proposed GEE. 

\subsection{Evaluation of GEE}

We evaluate two critical aspects of GEE: error coverage and explanation quality. 

\noindent \textbf{Error coverage evaluation} can be facilitated by forcing a model to generate position information of explained errors or to describe the edits being done. We thus conduct the evaluation by measuring (1) whether an explained error is indeed an error in the source and being corrected in the target; and (2) whether an error that is corrected in the target has an associated explanation.\footnote{A GEE model should be able to ignore errors in the source sentence that are not corrected in the target sentence since GEC is not its primary task.} An automatic evaluation through string overlap can give a quick estimate of error coverage when gold references are available. We also do manual evaluation to better understand the behavior of models.
% \kkcomment{mention how you do automatic eval? do you use string overlap vs gold data?}

\vspace{0.1in}

\noindent \textbf{Explanation quality evaluation} is challenging because errors can be explained in multiple ways. To reliably evaluate GEE outputs automatically, multi-reference metrics such as METEOR \citep{banerjee-lavie-2005-meteor} and benchmarks with multiple references for each error are needed. However, collecting such datasets is costly and requires expertise in second language teaching. Without such datasets being available, human experts are the only reliable evaluation. In our work, we recruit language teachers for evaluation described in \sectionref{sec:de_expl}. Language teachers, with their expertise in second language teaching, can reliably judge whether an explanation is correct and informative. 

% \micomment{this para doesnt say what you actually do to evaluate explanation quality. make it clear that you hire human experts and forward point to where you specify the eval task}

\section{Has \gpt\ already solved GEE?}\label{sec:gpt4_cant_end2end}

A natural question one might ask is whether  state-of-the-art LLMs can solve the GEE task in an end-to-end manner. In this section we demonstrate that \gpt~in its current form is error-prone. It has low error coverage and frequent hallucinations. Based on this observation, we experiment with an approach which provides \gpt~a list of manually extracted gold atomic edits. Results show that the edit list improves the performance greatly, indicating substantial headroom with more structured prompting (as we describe in \sectionref{sec:pipeline}).
\vspace{0.1in}

% Using these edits with the same naive one-shot prompt, the error coverage F1 is boosted from $0.64$ to $0.84$. This establishes the foundation for the pipeline that will be proposed in \sectionref{sec:pipeline}.
% \kkcomment{move these last three sentences to end of section? Or compress into one sentence saying something like, "additionally, we measure the headroom for performance improvement using manually labeled atomic edits, and notice improvements in performance which provides the foundation for our pipeline"}

% \kgcomment{i'd suggest thinking about what you really need this section to do. i think it's doing multiple things currently. it's preempting the objection ``can't GPT4 do this already?'' but also showing a sketch of the problem you're trying to solve and motivating the use of atomic edits.. but all three of these are latent rather than explicit aims of this section. would be easier for readers to follow if you set up the goal of the section at the outset, with both the section title and opening sentences. e.g., the section title or first sentence could be something like ``Is GEE a solved problem?''}

% \kgcomment{build up to the following claims with some more  setup. currently there's a gap from what came earlier in the paper to the following claims.}

% Although \gpt~is the most versatile state-of-the-art language model, it cannot do the GEE task in an end-to-end manner. It is very error-prone in the sense of low error coverage and frequent hallucinations. 

\noindent \textbf{One-shot prompting of \gpt}. We run an experiment using German grammar error correction data (details in \sectionref{sec:datasets}). We randomly sample 30 data points\footnote{Five data points from each CEFR level. Details are in \sectionref{sec:datasets}.} and generate explanations using the following one-shot prompt:

% . Among the 120 edits, there were 56 true positives, 27 false positives, and 37 false negatives

\begin{lstlisting}
You are given a pair of German sentences. The first sentence contains one or more errors, which are corrected in the second one. Your task is to: (1) generate a succinct explanation for each error following the template; (2) assign the error a type.

Template: The word X is deleted/inserted/replaced by Y/relocated because ...

Example:
Ich habe zwei bananen für mein Katze gekauft.
Ich habe zwei Bananen für meine Katze gekauft.
Explanation:
The word 'bananen' is replaced by 'Bananen' because German nouns should be capitalized.
Error type: capitalization
The word 'mein' is replaced by 'meine' because it should agree with the gender and case of the word Katze, which is feminine and accusative.
Error type: gender and case agreement

Below is the sentence pair for you to work on. Start with the explanation directly.
{src}
{trg}
Explanation:
\end{lstlisting}

% \micomment{where are the results? are you referencing some table?}

% Qualitatively, \gpt~makes mistakes even when sentences are short. As the example in \tableref{tab:base_gee_examples} shows, \gpt~hallucinates that \textit{Sie} is relocated. It also groups two errors together and omit the explanation of one error.
% % \kkcomment{move qualitative paragraph after quantitative}

% \input{sections/tables/base_gee_example}

\noindent \textbf{GPT-4 struggles to identify and explain errors.} To get a better sense of what mistakes are made by \gpt, we manually classify each edit as true positive, false positive, or false negative, depending on whether they are correctly mentioned in the generated explanations.\footnote{\textit{Correctly mentioned} means whether one can arrive from the source to the target through the edits.} Our evaluation shows that \gpt~mentioned 120 edits (compared to 90 edits by manual annotation), achieving recall $0.602$, precision $0.675$ and F1 score $0.636$. Such outputs are not useful for language learners as they wrongly consider correct words as errors. 
% \kkcomment{add bold mark headers for each paragraph. I've added some but feel free to change them}

\noindent \textbf{\gpt~makes mistakes even in short sentences.} As the example in \tableref{tab:base_gee_examples} shows, \gpt~hallucinates that \textit{Sie} is relocated. It also groups two errors together and omit the explanation of one error. The unorganized output format makes it hard to track the errors and follow the explanations. 
% \kkcomment{move qualitative paragraph after quantitative}

\begin{table}[h]
\fontsize{8}{10}\selectfont
\centering
\resizebox{\columnwidth}{!}{%
\begin{tabular}{p{2.5in}}
\toprule
\textbf{Source}: Bitte antworten sreiben Sie?       \\
\textbf{Target}: Bitte antworten und schreiben Sie. \\ \midrule
The word `sreiben' is replaced by `schreiben' because there was a spelling mistake in the word. \\
\textbf{Error type}: spelling                       \\ 
\colorbox{redish}{The word `Sie' is relocated after `antworten'} and the word `und' is inserted between `antworten' and `schreiben' because these are separate actions and should be connected with a conjunction. \\
\textbf{Error type}: \colorbox{redish}{word order} and conjunction     \\ \bottomrule
\end{tabular}}
\caption{An example of \gpt's GEE output with one-shot prompt. The second explanation hallucinates that `Sie' is relocated and does not give an explanation why it is relocated. The explanation also groups the relocation of `Sie' with the insertion of `und'.}
\label{tab:base_gee_examples}
\end{table}

\noindent \textbf{What if GPT-4 was provided with gold edits in prompt?} To measure the headroom for improvement, we prompt \gpt~using the same prompt but provide manually extracted gold atomic edits in the input prompt. Here, the recall, precision, and F1 are increased to $0.824$, $0.862$, and $0.843$ respectively. Hence, offering a good list of atomic edits to \gpt~is an important step of the process. This observation motivates our proposed pipeline in \sectionref{sec:pipeline}, where we augment \gpt~prompts with automatically extracted atomic edits.

\section{Pipeline for generating GEE}\label{sec:pipeline}

% \kkcomment{add one sentence connecting to previous section. Something like "in previous section we noticed that GPT-4 is bad at the task one-shot, but there is a lot of headroom for improvement using edit data in the prompt. Motivated by these findings,"} 

In \sectionref{sec:gpt4_cant_end2end}, we observed that adding a list of edits as the input to \gpt~can greatly improve the performance on error coverage. Motivated by the finding, we propose a two-step pipeline for GEE which uses atomic edit extraction as the intermediate step. The pipeline is illustrated in \figureref{fig:pipeline}. Given an input defined in \sectionref{sec:task.def}, we first extract atomic edits from the pairs. The edits are then appended to the sentence pair to form the input for the final step. In the last step, \gpt~is prompted to generate an explanation and an error type. 

% \kkcomment{this section generally needs more bold marked paragraph headings}

\subsection{Atomic edit extraction}

As discussed in \sectionref{sec:task.def}, we define an atomic edit as the smallest individual modifications at the token level, whose boundaries are decided linguistically. Each edit belongs to one of the four operation-level types: \texttt{replace}, \texttt{insert}, \texttt{delete}, and \texttt{relocate}. 
% \kkcomment{this paragraph should be first paragraph of this section. ERRANT one after that}
\vspace{0.1in}

\noindent \textbf{Previous work on edit extraction} ERRANT \citet{bryant-etal-2017-automatic} approaches the task in a rule-based linguistic manner. It extracts errors and label the errors with an edit type (e.g., deletion) and linguistic category (e.g., adverb). However, ERRANT has its limitations. For example, it does not account for relocated words.\footnote{It does account for local transposition (e.g., \textit{juice apple} vs.\ \textit{apple juice}).} It is also only designed for English. Adapting it to other languages requires great effort for individual languages \citep{korre-etal-2021-elerrant,uz-eryigit-2023-towards}. Further restrictions of ERRANT are discussed in \appendixref{appendix:errant}. As a result, we decide to use LLMs for atomic edit extraction. 

\vspace{0.1in}

% As discussed in \sectionref{sec:task.def}, we define an atomic edit as the smallest individual modifications made to a piece of text at the token level, whose boundaries are decided linguistically. Each edit belongs to one of the four operation-level types: replace, insert, delete, and relocate. \kkcomment{this paragraph should be first paragraph of this section. ERRANT one after that}

% \kgcomment{relocation too?}
% \kgcomment{so with this definition, you are preventing an atomic edit from being an edit to a character within a token? would be good to clarify and motivate that decision, as some people would naturally think of the smallest individual modification as changing one character of a token}

% \kgcomment{i think we should acknowledge and be clear about the subjective nature of defining atomic edits.. it is not merely about calculating a diff but is rather a task that requires knowledge and judgment to distinguish one atomic edit from another and can require semantic interpretation of the text and/or the writer's mental state. when errors are isolated in a sentence, maybe it's not too difficult to do, but when errors are consecutive in a sentence, there is sometimes a judgment call needed and experts may disagree about the atomic edit boundaries. would be nice to give an example of this in english so that readers can get the sense of it. i can help with an example if you need one.}

% \kkcomment{Maybe re-name paragraph heading below to "Desired LLM output format"?}
\noindent \textbf{Desired LLM output format.} To facilitate the evaluation of edit extraction and the GEE generation in the second step, we restrict the outputs of atomic edit extraction to a template \texttt{[operation type, original token(s), target token(s)]}. An example with all four edit types is given in (\ref{error_span}). 
% For a relocated token, the token should be the same before and after relocation, and the second and third slots should be the same. \kkcomment{can cut last sentence, obvious from example}

\begin{exe}
    \setlength\itemsep{0.2mm}
    \ex \label{error_span}
    m\"ochte \textbf{machen} \textbf{ein} Termine.\textbf{?}
    
    \textbf{Ich} m\"ochte \textbf{einen} Termine \textbf{machen}.

    \texttt{[insert, , Ich]}

    \texttt{[relocate, machen, machen]}
    
    \texttt{[replace, ein, einen]}

    \texttt{[delete, ?, ]}
\end{exe}

While being useful, the introduction of relocation occasionally reduces the model performance because models tends to label a relocated token as deletion plus insertion. Relocation can also be challenging for human to decide because a relocated word should be a word order error but have the same dependency in a sentence before and after relocation. We discuss details in \appendixref{appendix:manual_extract}.
\vspace{0.1in}

\noindent \textbf{Atomic edit extraction with LLMs.} To build an atomic edit extractor, we choose to prompt \claude,\footnote{\url{anthropic.com/index/introducing-claude}} \llama~\citep{touvron2023llama}, \chatgpt-0613, and Azure GPT-4 (2023-03-15-preview), as well as fine-tune \llama~and \chatgpt. For prompting, we use the carefully designed few-shot prompts in \appendixref{appendix:extraction_prompt} for German and Chinese. For fine-tuning, we use \llama~and \chatgpt~as the base models. We noticed that the models have a low recall when only sentence pairs are provided. To improve on that, we split sentences into a list of tokens and then extract rough string-based edits which are the longest contiguous matching subsequences.\footnote{We use Spacy for German and \href{https://github.com/fxsjy/jieba}{Jieba} for Chinese.} These rough edits are appended to sentence pairs as inputs. For all models, prompted or fine-tuned, we set temperature to $0$ because the task does not require creativity and temperature $0$ returns better performance compared to $0.2$ for all models.

\subsection{GEE generation}

\begin{figure}[t]
    \centering
    \includegraphics[width=0.48\textwidth]{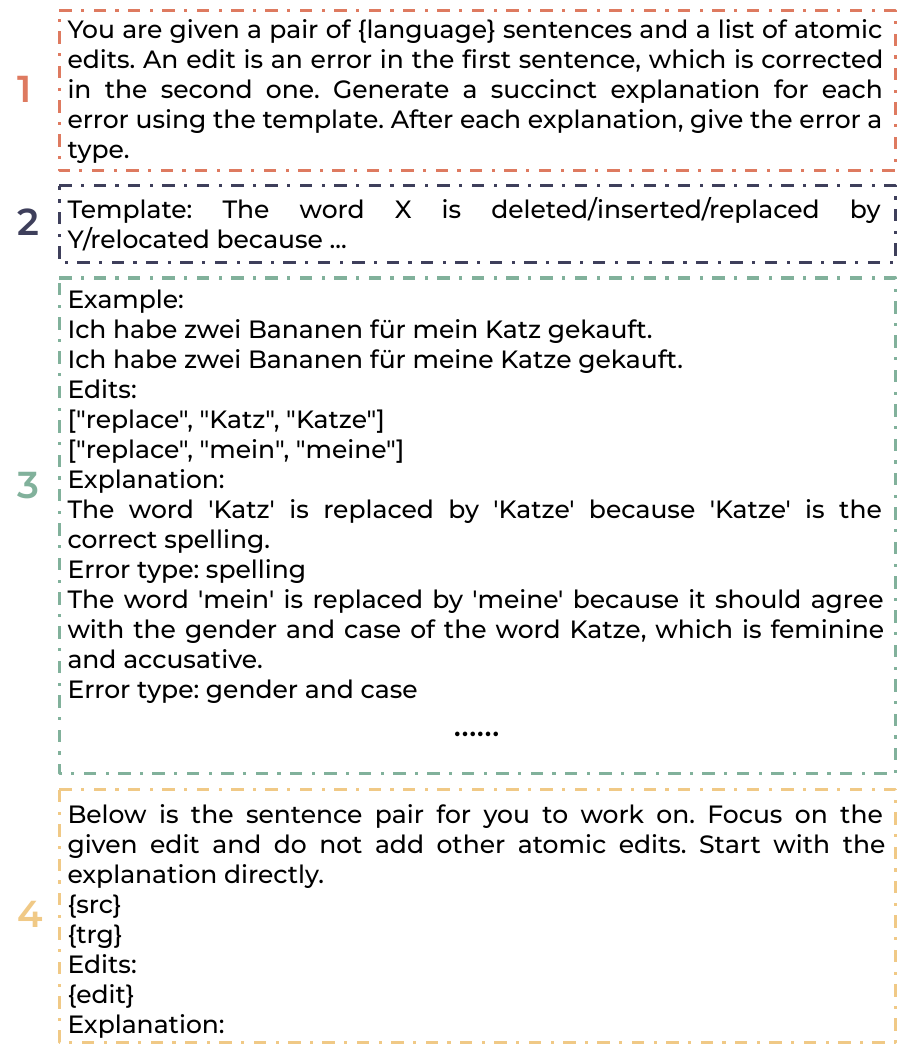}
    \caption{The prompt used for generating German grammar error explanation given an input defined in \sectionref{sec:task.def}. The prompt consists of: (1) task description, (2) generic explanation template, (3) few-shot examples, and (4) current input. The full prompts for German and Chinese are in \appendixref{appendix:expl.gen.prompts}.}
    \label{fig:expl.gen.prompt}
    \vspace{-0.15in}
\end{figure}

With the atomic edits being extracted, we are ready to generate explanations by prompting \gpt~using the template in \figureref{fig:expl.gen.prompt}. Edits are appended to the sentence pairs as the input. Given that each sentence pair may contain multiple errors, we investigated whether generating explanations for one error at a time or all explanations simultaneously would yield better results. In the prompt designing stage, we observed no significant difference in performance between the two approaches. Hence, we choose the latter strategy as it is efficient and cost-effective. 
% \kkcomment{give a high level overview of the approach --- say you will use GPT-4 prompting with the atomic edits you extracted from previous section, and then point to Fig 2}

% \kkcomment{add a bridge sentence reminding readers that sentence pairs may have multiple edits} 

\figureref{fig:expl.gen.prompt} gives a shortened example of the German GEE prompt. The full prompts for German and Chinese are  in \appendixref{appendix:expl.gen.prompts}. The prompts consists of four parts. The first part is the \textbf{task description}, which is followed by a generic \textbf{template} of explanations. Below the template are few-shot \textbf{examples}. In the examples, we aim to offer both meta-linguistic and meaning-oriented explanations whenever it is possible as they help L2 users improve their language skills (i.e., using languages accurately and fluently) \citep{lyster2010interactional}. At the end of the prompt, we provide \gpt~the \textbf{sentence pair with a list of atomic edits} and ask the model to generate one explanation with an error type for each edit. The generated outputs have the following format:

\begin{lstlisting}
[edit description] because [edit reason]
Error type: [error type]
\end{lstlisting}

\noindent The edit description describes how a word in the source sentence is edited in the target sentence. The edit reason explains why such an edit is made.

\section{German and Chinese datasets}\label{sec:datasets}

% \kgcomment{feels a bit early to discuss dataset details and data processing. that would be more appropriate for an experimental setup section. i think you want to get to the methods sooner and only mention dataset info in those methods sections that is necessary for readers to understand the methods}

This section introduces the datasets that are used in our experiments. Statistics of the sampled data subsets are reported in \tableref{tab:dataset.pair.edit.count}.

\begin{table}[h]
\fontsize{8}{10}\selectfont
\centering
\resizebox{\columnwidth}{!}{%
\begin{tabular}{rrrrr}
\toprule
& \multicolumn{2}{c}{German} & \multicolumn{2}{c}{Chinese} \\
& \# of data   &   \# of edits      & \# of data  &   \# of edits \\\midrule

Fine-tune & 500    & 1598  & 496   & 790 \\            
Test      & 50     & 186   & 53    & 94  \\
GEE       & 1122   & --    & 970   & --  \\ \bottomrule
\end{tabular}}
\caption{Number of sentence pairs and gold edits in each data subset in German and Chinese. We do not manually annotate the data for GEE, hence no gold edit count is reported.}
\label{tab:dataset.pair.edit.count}
\end{table}

\subsection{German Merlin and Falko}

For German GEE, we use the data from the German L2 learner corpora Falko EssayL1v2.3 \citep{ludeling2008lernerkorpus, reznicek2010falko} and Merlin \citep{boyd-etal-2014-merlin}. Both datasets consist of essays written by German learners whose proficiency ranges from beginners to advanced users. The datasets provide corrections of errors. The datasets are preprocessed as described in \appendixref{appendix:data.preprocess}.

% proficiency levels A1 to C1 according to the Common European Framework of Reference for Languages (CEFR).\footnote{\href{https://www.coe.int/en/web/common-european-framework-reference-languages}{The Common European Framework of Reference for Language} (CEFR) is a standard for describing language ability. There are six levels: A1, A2, B1, B2, C1, and C2. C2 is the native speaker level.} The Merlin dataset is a collection of essays written by advanced German speakers from different countries with both native and non-native background. We use the data from Merlin as C2 data. The datasets are preprocessed as described in \appendixref{appendix:data.preprocess}.

% Both Falko and Merlin offer two types of grammar error corrections, target hypothesis 1 and target hypothesis 2. Target hypothesis 1 performs minimal correction at the morpho-syntactic level while target hypothesis 2 modifies semantic and pragmatic aspects (e.g., information structure or word choice) of the input text, aiming for a more advanced paraphrase-type correction. For our purpose, we use target hypothesis 1 of each corrected sentence.\footnote{Examples of the target hypothesis 1 and 2 of a corrected sentence can be found in \url{https://gucorpling.org/amir/pdf/Reznicek_et_al.pdf}.}

% To prepare the datasets, we first split the paragraphs in Falko and Merlin into sentences by adapting the paragraph alignment algorithm in \citet{thai-etal-2022-exploring} for sentence alignment. We then screened out sentence pairs that: (1) have short sentences (less that 3 tokens); (2) contain ``incomp'' or ``unreadable'' tokens; and (3) have two sentences in the source and one sentence in the target, or vice versa. 

From the preprocessed dataset, we sample two subsets without overlaps between them. 
% First, we sample 200 data points from each level from A2 to C2 for fine-tuning and testing atomic edit extraction models. Out of the total 1000 data points, we manually annotated $550$ data points and use $500$ for fine-tuning and $50$ for testing, each containing $1598$ and $186$ gold edits. 
First, we sample $550$ data points and manually annotate them for gold atomic edits. The $550$ data points are split into $500$ for fine-tuning and $50$ for testing, each containing $1598$ and $186$ gold edits. 
Second, for GEE generation, we sample all A1 data points (146) and randomly sample 200 data points from other CEFR levels (A2--C2). We manually remove sentence pairs that are misaligned.\footnote{An example 
%for misaligned sentence pairs 
is \textit{vielen Dank für die Einladung.} and \textit{12.3.2012 Liebe Silke, vielen Dank für die Einladung.}} At the end, we have 1122 sentence pairs in German for GEE.

\subsection{Chinese CGED2017}

We conduct the Chinese GEE experiment on the training split of Chinese Grammatical Error Diagnosis (CGED) 2017 \citep{rao-etal-2020-overview}, which are from the writing task of the \textit{Hanyu Shuiping Kaoshi} (Test of Chinese Level) \citep{cui2011principles, zhang2013design}. Error corrections are provided but there is no learner proficiency level information. 

Data are preprocessed as in \appendixref{appendix:zh.data.preprocess}. We sampled 520 and 60 data points for fine-tuning/prompting edit extraction models and testing performance respectively. We sample another 970 data points for generating error explanations. After cleaning, we have 496 data points for fine-tuning, 53 for testing, and 970 for explanation generation. Edit counts are in \tableref{tab:dataset.pair.edit.count}.

% \section{Atomic Edit Extraction Results}\label{sec:edit.extract.results}
% \kgcomment{the German dataset details could be moved to this section}
\section{Experimental results}\label{sec:edit.extract.results}

% \kkcomment{organize this as follows: Section 6: Experimental Results. Section 6.1: Atomic Edit Extraction. Section 6.1.1: AEE German. Section 6.1.2: AEE Chinese. Section 6.2: GEE. Section 6.2.1: GEE German. Section 6.2.2 GEE Chinese. Add high level takeaways for each section/subsection}

This section presents the results of the GEE pipeline in German and Chinese. We first present the results of the fine-tuned and prompted models on atomic token edit extraction in \sectionref{sec:results.on.aee}. We find that the fine-tuned \chatgpt~achieved the best performance on edit extraction for German but \gpt~works the best for Chinese. \sectionref{sec:human_eval} presents the human evaluation results of German and Chinese GEE outputs generated by \gpt. Our human evaluation shows promising performance. Among the German GEE outputs, $93.9\%$ are judged by two German teachers as correct. For Chinese GEE outputs, $98\%$ of the outputs are correct according to a Chinese teacher.

\subsection{Atomic edit extraction results}\label{sec:results.on.aee}

We first introduce the experiment and evaluation setup, then measure the performance of fine-tuned and prompted models with respect to recall, precision, and F1. Results are presented in Tables~\ref{tab:de_edit_accuracy} and \ref{tab:zh_edit_accuracy} for German and Chinese,  respectively. In general, all models follow the output format well but the fine-tuned \chatgpt\ performs best for German and \gpt\ is best for Chinese.

% \subsubsection{German Atomic Edit Extractor}

% We first present the results of German atomic edit extraction. 

% \kkcomment{clearly distinguish what's an experimental setup detail and what's a result.}

\noindent \textbf{Experiment setup.} We few-shot prompt \claude, \chatgpt, and \gpt~with the prompt for German in \appendixref{appendix:de_extraction_prompt}.\footnote{We also prompted \llama~in the same way but its F1 is only 0.086. The outputs contain repetition and irrelevant content which makes them unusable.} For fine-tuning, we use \llama~and \chatgpt~as the base models and fine-tune them on the 500 training data points in \tableref{tab:dataset.pair.edit.count}. Details of the fine-tuning process are in \appendixref{appendix:fine-tune.detail}. At inference time, the temperature of all models is set to $0$. We employ simple heuristics to post-process model outputs to remove low-level false positive errors, such as replacement edits that have the same original and edited tokens.
% We use the test data reported in \tableref{tab:dataset.pair.edit.count}. 

\noindent \textbf{Evaluation.} While automatic evaluation is fast, we evaluate the test data manually because there can be multiple ways to get to a target sentence from a source sentence. Concretely, we compare model edits against the manually extracted gold edits one by one. When there is a discrepancy, if the model outputs are linguistically meaningful and can reach the same target, we treat them as true positives. 

% \kkcomment{make paragraph heading below more informative. Something like "Results: fine-tuned GPT3.5 is most effective at atomic edit extraction"}
% \noindent \textbf{German extractor results} 
\noindent \textbf{Results on German: fine-tuned GPT3.5 is most effective at atomic edit extraction.} The results for German edit extraction in terms of precision, recall, and F1 are in \tableref{tab:de_edit_accuracy}. All models have reasonable performance but the fine-tuned \chatgpt\ outperforms all others. It achieves $0.923$ in recall, $0.939$ in precision, and $0.931$ in F1. We use it as the atomic edit extractor in the next step in German GEE generation.

% \begin{table}[]
% \centering
% \resizebox{0.98\columnwidth}{!}{%
% \begin{tabular}{@{}rrrrrrr@{}}
% \toprule
%            & \multicolumn{1}{c}{\claude} & \multicolumn{2}{c}{\llama} & \multicolumn{2}{c}{\chatgpt} & \multicolumn{1}{c}{\gpt}  \\
%            & Prompting     & Prompting     & Fine-Tuned    & Prompting    & Fine-Tuned    & Prompting        \\ \midrule
% Recall     & 0.789         & 0.116         & 0.849         & 0.695        & \textbf{0.923}& 0.874           \\
% Precision  & 0.737         & 0.069         & 0.827         & 0.764        & \textbf{0.939}& 0.870           \\
% F1         & 0.762         & 0.086         & 0.838         & 0.728        & \textbf{0.931}& 0.870           \\
% Edit Count & 199           & 321           & 191           & 161          & 180           & 184             \\ \bottomrule
% \end{tabular}}
% \caption{The recall, precision, and F1 scores of the models on German atomic edit extraction task. Temperature of all models is set to $0$ at the inference time. Because of the variance in \gpt~outputs, the outputs are generated three times and the best performance is reported.}
% \label{tab:de_edit_accuracy}
% \end{table}
% % results for GPT3.5 fine-tuned rule-screened temp = 0 https://docs.google.com/spreadsheets/d/1lWVZ43xxunqJ6n5lNMjXCD6pCrfzM0jdQMWzyGS6riQ/edit?usp=sharing

\begin{table}[]
\centering
\resizebox{0.98\columnwidth}{!}{%
\begin{tabular}{@{}rrrrrrr@{}}
\toprule
           & \multicolumn{1}{c}{\claude} & \multicolumn{1}{c}{\llama} & \multicolumn{2}{c}{\chatgpt} & \multicolumn{1}{c}{\gpt}  \\
           & Prompting     & Fine-Tuned    & Prompting    & Fine-Tuned    & Prompting        \\ \midrule
Recall     & 0.789         & 0.849         & 0.695        & \textbf{0.923}& 0.874           \\
Precision  & 0.737         & 0.827         & 0.764        & \textbf{0.939}& 0.870           \\
F1         & 0.762         & 0.838         & 0.728        & \textbf{0.931}& 0.870           \\
Edit Count & 199           & 191           & 161          & 180           & 184             \\ \bottomrule
\end{tabular}}
\caption{Recall, precision, and F1 scores of models on the German atomic edit extraction task. 
%Temperature of all models is set to $0$ at inference time. \kgcomment{i commented out the sentence here about temperature since it's already stated in the main text and not super important}
Because of the variance in \gpt\ outputs, the outputs are generated three times and the best performance is reported.}
\label{tab:de_edit_accuracy}
\end{table}
% results for GPT3.5 fine-tuned rule-screened temp = 0 https://docs.google.com/spreadsheets/d/1lWVZ43xxunqJ6n5lNMjXCD6pCrfzM0jdQMWzyGS6riQ/edit?usp=sharing

% \subsection{Chinese Atomic Edit Extractor}

\noindent \textbf{Results on Chinese: prompted \gpt~is the most effective edit extractor.} 
%For the Chinese atomic edit extractor, we experiment with the four models in the same way as in German.
% \footnote{We do not prompt \llama~because it does not follow the format in the instruction and generates repetition or irrelevant content.} 
The results are %The model recall, precision, and F1 scores are 
reported in \tableref{tab:zh_edit_accuracy}. Unlike German, the prompted \gpt\ returns the best performance. Because of the variance in the \gpt\ outputs, we verify its performance by running the experiment three times. All three runs of \gpt\ return the highest scores. The best results of \gpt\ are recall 0.884, precision 0.933, and F1 score 0.908. We hypothesize that the reason of the prompted \gpt~performing well on Chinese is that each Chinese sentence pair has less edits on average (see \tableref{tab:dataset.pair.edit.count}). The same reason leads to the fact that there are less edits in the training data, which might cause the fine-tuned models perform worse than the ones in German. 
% \kgcomment{i did some shortening here. we can shorten further. but are there any other things we can say about the results? e.g., any speculation as to why GPT-4 is best for Chinese while finetuned 3.5-turbo is best for German?}
% \kkcomment{Move the result sentences to new paragraph. Specify clearly that this part is talking about results. Add more sentences to it + copy numbers from the table in-line}

\begin{table}[]
\centering
\resizebox{\columnwidth}{!}{%
\begin{tabular}{@{}rrrrrr@{}}
\toprule
           & \multicolumn{1}{c}{\claude} & \multicolumn{1}{c}{\llama} & \multicolumn{2}{c}{\chatgpt} & \multicolumn{1}{c}{\gpt} \\
           & Prompting     & Fine-Tuned    & Prompting     & Fine-Tuned         & Prompting                       \\ \midrule
Recall     & 0.872         & 0.840         & 0.763         & 0.830              & \textbf{0.884}           \\
Precision  & 0.820         & 0.908         & 0.651         & 0.918              & \textbf{0.933}            \\
F1         & 0.845         & 0.873         & 0.703         & 0.872              & \textbf{0.908}\\
Edit Count & 100           & 87            & 109           & 85                 & 90                 \\ \bottomrule
\end{tabular}}
\caption{Recall, precision, and F1 scores of models in the Chinese atomic edit extraction task. 
%Temperature of all models is set to $0$ at the inference time. \kgcomment{i commented out the sentence here about temperature since it's already stated in the main text and not super important}
Because of the variance in \gpt\ outputs, the outputs are generated three times and the best performance is reported.}
\label{tab:zh_edit_accuracy}
\end{table}

\subsection{Human evaluation of GEE}\label{sec:human_eval}

% This section provides quantitative results from the human evaluations of \gpt~on the generated GEE for German and Chinese. Detailed qualitative analysis is in \appendixref{appendix:de.expl.quality.analysis}. 

% \footnote{One teacher gives classes 15 hours a week to students whose level ranges from A1 to C2. The other teacher teaches students from A1 to C1 20 hours per week.}

To evaluate the performance of our GEE pipeline, we recruited two German teachers and one Chinese teacher.\footnote{Both German teachers give classes 15 to 20 hours per week. The Chinese teacher teaches 4 classes a week.} This section provides quantitative results from the human evaluations of \gpt~on the generated GEEs for German and Chinese. Detailed qualitative analysis is in \appendixref{appendix:de.expl.quality.analysis}. 

The results indicate that our GEE pipeline generates $93.9\%$ and $98\%$ correct explanations for German and Chinese, respectively. However, we find that \gpt~occasionally produces low-level errors such as formatting issues. For Chinese, when it comes to word choice errors, \gpt~does not always provide clear contrast between two words. 

% \subsection{German GEE Generation}
\subsubsection{Human evaluation of German GEE}\label{sec:de_expl}

\noindent \textbf{German GEE generation.} Using the best performing edit extractor from \sectionref{sec:results.on.aee}, we extract atomic edits from the $1122$ sentence pairs described in \sectionref{sec:datasets}.
% and get $3336$ edits in total. 
The extracted edits are paired with the source and target sentences to prompt \gpt~using the few-shot prompt in \appendixref{appendix:de.expl.gen.prompts}. We use the default hyperparameters offered by the OpenAI API (i.e., temperature $= 1$ and top p $= 1$) for some creativity in the explanations. 

% \kkcomment{Section too short, merge with next section}
% The generated outputs have the following format:

% \begin{lstlisting}
% [edit description] because [edit reason]
% Error type: [error type]
% \end{lstlisting}

% \noindent The edit description describes how a word in the source sentence is edited in the target sentence. The edit reason explains why such an edit is made. The model is also required to assign an error type to the explained error. 

% \subsection{}

% \section{Human Evaluation of German GEE}\label{sec:de_expl}

% \kkcomment{merge German and Chinese human GEE eval into one section. Again use paragraph headers and clearly specify what's experimental setup details and what's a result}

\vspace{0.1in}

\noindent \textbf{German GEE evaluation setting.} The annotation interface is shown in \figureref{fig:annotation_interface}. We collected annotations on error explanations of 596 unique German sentence pairs. To assess the agreement between the teachers, 96 pairs are annotated by both of them. A total of 692 sentence pairs were annotated for this study.\footnote{There are 2082 edits extracted from 692 sentence pairs, but \gpt\ only generates explanations for 1986 of them.} The two teachers' agreement rate is $89.6\%$. Details of the agreement assessment and evaluation instructions are in \appendixref{appendix:human.eval}.

\vspace{0.1in}

\noindent \textbf{Human annotation protocol for evaluating GEE.} For each sentence pair, we present the explanations generated by \gpt~to the teachers, who are asked to check for four types of mistakes in the explanations:\footnote{We call grammar errors in sentences as errors and errors made by \gpt~as mistakes.} 
% \kkcomment{break error types into bullet points}

\begin{itemize}[leftmargin=*]
\setlength\itemsep{0mm}
    \item \textbf{Hallucinated error}: an error in an explanation that does not exist in the source sentence. Such a mistake can be made by considering a correct word/punctuation as an error, or it can be a word that does not exist in the sentences at all. 
    
    \item \textbf{Missing error}: a true error in the source sentence, which is edited in the target sentence but not explained.

    \item \textbf{Wrong error explanation}: wrong edit description, wrong edit reason, or both.

    \item \textbf{Wrong error type}: an error type that is not related to the explained error.
\end{itemize}

\begin{table}[]
\centering
\resizebox{0.85\columnwidth}{!}{%
\begin{tabular}{@{}lrr@{}}
\toprule
                        & Count & Percentage \\\midrule
Fully correct           & 1865  & 93.9\%    \\
Wrong edit description  & 65    & 3.3\%     \\
Wrong edit reason       & 29    & 1.5\%     \\
Wrong error type        & 12    & 0.6\%     \\
Hallucinated error      & 15    & 0.8\%     \\\midrule
Total explanation count & 1986  & 100\%      \\
Total annotated items   & 692   &            \\
Missing error           & 67    &    \\\bottomrule
\end{tabular}}
\caption{
% \kkcomment{Caption still needs work. Bold mark the task being evaluated (German GEE) along with human evaluation. Add high-level results takeaway from the  table. Move details like number of data points to third sentence.}
Results of human evaluation on German GEE by two German teachers. 692 sentence pairs with 1986 explanations are annotated. \gpt~generates fully correct edit description, edit reason, and error type $93.9\%$ of the time. Low-level \textit{wrong edit descriptions} count for $3.3\%$ of the mistakes. The count of \textit{missing errors} by the teachers is the lower bound of the actual ones.}
\label{tab:german.mistake.count}
\end{table}
% python code/German/human_eval/result_analysis/all_anno_excel_prep.py 

% \kkcomment{add paragraph heading with overall takeaway}

\noindent \textbf{German GEE using edit-driven GPT-4 prompts has high quality.}
The counts of each mistake type are reported in \tableref{tab:german.mistake.count}. The results show that \gpt~generates correct explanations $93.9\%$ of the time. The occurrences of inappropriate error types and hallucinated errors are both below $1\%$. Among the 94 \textit{wrong error explanations}, 65 are wrong in the edit description but correct in edit reason. As many as 31 edit description mistakes are made because \gpt\ describes inserted and deleted edits as \textit{The word `' is inserted/deleted because ...} without mentioning the word itself. Among the 15 hallucinated errors, 12 are caused by wrong atomic edit extraction and 2 are hallucinated by \gpt\ in the process of generating explanations.\footnote{One annotated item accidentally has the same source and target sentences. Its atomic edit list is empty but \gpt\ hallucinates that there is an error in the source sentence that is corrected in the target sentence.} 

\noindent \textbf{Remaining issues.} To gain a deeper understanding of \gpt's limitations, we look into its mistakes in detail and notice that \gpt\ does not consider a context that is sufficiently large for certain errors, especially when it comes to prepositions. For example, when explaining the error in \textit{mit 2 Zimmer} vs.\ \textit{mit 2 Zimmern}, \gpt\ only says that the dative case is needed here. It does not consider the close-by preposition \textit{mit} which requires a dative case of its complement. We provide a detailed analysis of other errors in the \gpt~outputs in \appendixref{appendix:de.expl.quality.analysis}. 

% \kkcomment{Can you provide a table in the main body discussing the major themes of wins and losses with this model? Same for Chinese. Keep some qualitative analysis in the main body as well. 2 rows from Table 8 and 9 each can appear in main body IMO, it's ok if it takes up to half to one page here since this is the very end of paper}

\subsubsection{Human evaluation of Chinese GEE}

To understand how generalizable our pipeline is to different types of languages, we evaluate its performance on Chinese using the CGED2017 data described in \sectionref{sec:datasets}. One Chinese teacher evaluated Chinese GEE outputs on 200 sentence pairs with 302 explanations.\footnote{There are 310 edits extracted from these 200 sentence pairs. \gpt\ only generates explanations for 302 of them.} The annotation task is set up in the same way as German. 

% \kkcomment{this paragraph sounds negative even though the result is positive, I think it's because only one sentence is on the main result. Bold marked headers separating the positive / negative findings will help here.}

\begin{table}[h]
\centering
\resizebox{0.85\columnwidth}{!}{%
\begin{tabular}{@{}lrr@{}}
\toprule
                        & Count & Percentage \\\midrule
Fully correct           & 296  & 98.01\%    \\
Wrong edit description  & 1    & 0.03\%     \\
Wrong edit reason       & 3    & 0.10\%     \\
Wrong error type        & 2    & 0.07\%     \\
Hallucinated error      & 0    & 0.0\%     \\\midrule
Total explanation count & 302  & 100\%      \\
Total annotated items   & 200  &            \\
Missing error           & 0    &    \\\bottomrule
\end{tabular}}
\caption{
% \kkcomment{Caption still needs work. Bold mark the task being evaluated (Chinese GEE) along with human evaluation. Add high-level results takeaway from the  table. Move details like number of data points to third sentence.}
Results of human evaluation on Chinese GEE by one Chinese teacher. $98\%$ of the generated explanations are judged as correct. 200 sentence pairs with 302 explanations are annotated. The evaluation criteria are the same as for German.}
\label{tab:chinese.mistake.count}
\end{table}

\noindent \textbf{Positive findings.} Among the 302 annotated explanations, $98\%$ are judged as correct by the Chinese teacher. \gpt\ has very low mistake rates in all four mistake types. This shows that the proposed pipeline is effective and adaptable for very different languages like German and Chinese.

\noindent \textbf{Remaining issues.} While \gpt~achieves high correctness rate in Chinese GEE, there are two caveats. First, during the annotation of the data for gold atomic edits, we notice that most of the edits are simple and can be readily extracted by a string-based tool. One reason is that each sentence pair on average has fewer edits than in the German data (3.24 vs.\ 1.61, see \tableref{tab:dataset.pair.edit.count}). 
% Second, some atomic edits with multiple characters do not form a word or a phrase. Hence, they do not satisfy the requirement of defining error boundaries in a linguistically informed way. 
Second, for word choice errors, \gpt~does not always give a clear comparison of word meanings. 

\vspace{0mm}

\begin{exe}

    \ex\label{ex.word.choice.expl} 
    \zh{\colorbox{redish}{严重性}的问题} $\rightarrow$ \zh{\colorbox{redish}{严重}的问题}
    
    The word '\zh{严重性}' is replaced with '\zh{严重}' because '\zh{严重}' is the correct word for 'serious' when describing the severity of a problem.

\end{exe}

\vspace{0mm}

\noindent For example, in (\ref{ex.word.choice.expl}), \gpt\ explains what \zh{严重} (serious) means but it does not explain why \zh{严重} is good in \zh{严重的问题} (serious problem) but \zh{严重性} (seriousness) is not. Because word choice is a prevalent problem in Chinese grammar errors (see \tableref{tab:chinese.error.types} for error types generated by \gpt), such clear comparison should be enforced in an explanation so that language learners can draw inferences about other cases from the current error. 
% \kkcomment{Move example higher up? It will act as paragraph break and move example closer to reference}

\section{Related work}

% \kkcomment{categorize your related work. Add bold marked paragraph headings for each different theme. Themes could be "GEC", "GEC with edit types", "Explaining GEC errors"}
% \kkcomment{add overall bird's eye view paragraph here. Say something like "There have been several related works, the most prominent themes being XYZ. Below, we describe each of them in detail and describe how our work builds on them."}

Our GEE task is built upon the actively studied GEC task. The task is often formulated as a neural machine translation task, with the source being a piece of text with grammar errors and the target being the grammar-error-free text \citep{boyd-2018-using,bryant2022grammatical,yuan-bryant-2021-document,zhang-etal-2022-mucgec}. Researchers in the GEC domain have explored various aspects of the task. We identify two of them which the GEE task can be built on and benefit from. We also compare our task to a related task, feedback comment generation (FCG), and show how GEE is different from it.

\vspace{0.1in}

\noindent \textbf{GEC with multi-reference and context.} Research has been building GEC models on data which have one gold reference for each source input. However, there is an urge to use multiple references for source inputs \citep{bryant-ng-2015-far,zhang-etal-2022-mucgec,xu-etal-2022-fcgec}. In the context of GEE, a capable model should generate well-suited explanations for any valid error corrections, which requires reasoning of word relations and recovering correction rationales, not just memorize grammar rules. Such ability of GEE models also need to go beyond the sentence level. \citet{wang-etal-2022-cctc} has shown that even when only one sentence is added to the input as the context, a GEC model's performance can be significantly boosted. If some errors can only be better corrected in context, they can be better explained in context as well. 

\vspace{0.1in}

\noindent \textbf{GEC with auxiliary grammar information.} 
There are works that have shown improvement of GEC models by adding edit types, dependency information, or grammatical error type into the training process \citep{omelianchuk-etal-2020-gector, ma-etal-2022-linguistic, yang-etal-2023-leveraging}. \citet{fei-etal-2023-enhancing} study the influence of adding evidence words for errors and error types into the pipeline of GEC. They found that such information can significantly increase model performance in English GEC. For the GEE task, it is an interesting direction to explore whether adding those extra information to a GEE system can improve its explanations' usefulness. 

\vspace{0.1in}

\noindent \textbf{Feedback alongside grammar error detection.} 
On the side of explanation in GEC, \citet{nagata-etal-2021-shared} proposed a generative shared task called \textit{feedback comment generation for language learners} (FCG). It is based on the dataset ICNALE Learner Essays with Feedback Comments Dataset \citep{nagata-etal-2020-creating}. The task differs from our GEE task in three important aspects. First, the FCG task is built on the grammar error detection task which does not correct errors. The inputs in FCG are erroneous sentences only, which have spans marked as errors. Hence, the FCG task does not need to handle the problem of extracting errors in a linguistically informed way. Second, the FCG task focuses only on preposition words, which are a closed set of function words whose occurrences and usages are limited. In our task, the involved error types have a wide range, as listed in \tableref{tab:chinese.error.types} for German and \tableref{tab:german.error.types} for Chinese. Third, while the FCG task focuses on generating comments as hints for language learners to correct errors themselves,\footnote{An example comment given in \citet{nagata-etal-2021-shared} is \textit{``Agree'' requires a preposition since it is an <intransitive verb>. Look up the appropriate preposition in a dictionary.}} our task aims to enhance learners' knowledge by showing them the corrected sentences, the underlying grammar rules, and a comparison with errors and corrected words when necessary.

\vspace{0.1in}
% \kkcomment{add paragraph bold marked title here}
\noindent \textbf{Works on feedback comment generation}
\citet{coyne-2023-template} and \citet{coyne-2023-developing} work on the FCG task and develop a typology for learning feedback, including abstract types (e.g., tone and idiom) and grammatical pattern types (e.g., comparative and causative). However, their work is in an early stage with no human or automatic evaluation on the comment quality. \citet{behzad-etal-2023-sentence} present a strong baseline for the FCG task but points out that, at the current stage, many feedback comments are generic (e.g., \textit{Look up the use of the <verb> X in a dictionary and rewrite the sentence using the appropriate structure.}) \citet{stahl-wachsmuth-2023-identifying}, \citet{jimichi-etal-2023-feedback}, and \citet{ueda-komachi-2023-tmu} approach the FCG task via fine-tuning language models such as T5~\citep{raffel2020exploring} or BART~\citep{lewis2020bart}. However, for the GEE task, especially when there are restricted annotated resources for fine-tuning, it is unclear whether such an approach can work. Lastly, these works evaluate model outpus with BLEU~\cite{papineni2002bleu} and lack careful human evaluation.

\section{Conclusion}

We present a new task grammar error explanation, where systems provide natural language explanations to users explaining the grammatical errors they made. We find that \gpt~cannot perform this task with high accuracy using one-shot prompting, and hence develop a pipelined approach using LLMs and atomic token edits to generate grammar error explanations. We find that our LLM-based pipeline gets a high score of $93.9\%$ in German and $98\%$ in Chinese error explanation. 

While \gpt~achieve high correctness rate in Chinese error explanation, our Chinese teacher identified $28\%$ of the data that do not fully correct errors in source sentences. Further research is encouraged to build datasets that have the following two properties for GEE. First, the dataset should include data from all proficiency levels of language learners so that we can readily evaluate GEE systems' performance on a wide range of error types. Second, the dataset should provide high quality correction so that a GEE system can leverage context information when generating explanations. 

While we assume a grammar error correction system as the foundation of our GEE system, further work are encouraged to explore GEE generation alongside GEC. 

\section*{Limitations}

We acknowledge two limitations of our current work. First, our grammar error explanation system only considers sentence level inputs. However, certain error types (e.g., word choice and coreference) can benefit from a larger context. Second, because the Chinese data used in our work are from the HSK test (Test of Chinese Level), the covered topics are limited. It also does not include data from learners from all proficiency levels. Hence, the error types might not be representative for all levels of Chinese learners.

% \yscomment{Mohit: paper has many high-level organizational issues\\
% there needs to be a precise definition of the task after the intro\\
% currently it reads just like a list of what you did}

% \yscomment{
% all of these details are nice but not useful when the task and contributions are unclear\\
% intro\\
% define the task of GEE clearly\\
% methods for GEE: (1) prompt an LLM; (2) extract atomic edits and then generate explanation conditioned on the edits\\
% data collection and experimental setup for GEE\\
% results: (1) german; (2) chinese\\
% related work\\
% conclusion\\
% theres also in general just a lot of details that you could likely move to appendix to make the exposition more clear}

% \input{sections/ethics}

% \input{sections/acknowledgements}

\bibliography{anthology,custom}
\bibliographystyle{acl_natbib}

\appendix
\section{Reasons of not using ERRANT}\label{appendix:errant}

ERRANT \citep{bryant-etal-2017-automatic} is an effort to standardise datasets for GEC, reduce annotators' burden, and offer feedback to instructors and learners. It does so by offering a tool that automatically extracts and labels edits in the format of \texttt{operation:linguistic feature}. 

ERRANT would have been ideal for our purpose. Concretely, this would have been ideal for the edit extraction in Step 1 and error type tagging in Step 2. However, ERRANT has several shortcomings. 

First, ERRANT is designed only for English and its error type tagging process is based on a English rule-based framework. Extending it to another language will take great effort \citet{korre-etal-2021-elerrant, uz-eryigit-2023-towards}.

Second, there is ambiguity in ERRANT's error type names. For example, \texttt{R:ADV} is a possible error type in ERRANT in which \texttt{R} stands for replacement and \texttt{ADV} stands for adverb. But it is not clear, as it stands, whether it represents only an adverb being replaced by another adverb, or it could be the case that a word of other category is replaced by an adverb. 

Third, \citet{korre-pavlopoulos-2020-errant} show that ERRANT can falsely or ambiguously tag errors. In their work, they use ERRANT to tag the errors in the FCE dataset \citep{yannakoudakis-etal-2011-new}. They then sample 100 sentence pairs to whose errors ERRANT assigned the type \texttt{Other}. They examine those sentence pairs and found that up to $39\%$ of the data point could have been assigned a more precise label. 

Fourth, ERRANT's underlying edit extractor does not account for non-local token relocation \citep{felice-etal-2016-automatic}. The extractor aligns the tokens in the erroneous and correct sentences and assign one of the following labels to spans: \texttt{M(atch)}, \texttt{I(nsertion)}, \texttt{D(eletion)}, \texttt{S(ubstitution)}, and \texttt{T(ransposition)}. For a relatively locally relocated token, the extractor assigns the span as \texttt{T} as in \ref{transposition}. However, for a less local token relocation such as \ref{tok_relocation}, the extractor treats it as being deleted then inserted.

\begin{enumerate}[label=(\arabic*)]
    \item\label{transposition}
    Ich\textsubscript{0} m{\"o}chte\textsubscript{1} \textbf{haben}\textsubscript{2} einen\textsubscript{3} Apfel\textsubscript{4} .\textsubscript{5} 

    Ich\textsubscript{0} m{\"o}chte\textsubscript{1} einen\textsubscript{2} Apfel\textsubscript{3} \textbf{haben}\textsubscript{4} .\textsubscript{5}

    \texttt{(`M', 0, 1, 0, 1)}

    \texttt{(`M', 1, 2, 1, 2)}

    \textbf{\texttt{(`T3', 2, 5, 2, 5)}}

    \texttt{(`M', 5, 6, 5, 6)}
    
    \item\label{tok_relocation}
    Ich\textsubscript{0} m{\"o}chte\textsubscript{1} \textbf{haben}\textsubscript{2} einen\textsubscript{3} roten\textsubscript{4} Apfel\textsubscript{5} .\textsubscript{6} 

    Ich\textsubscript{0} m{\"o}chte\textsubscript{1} einen\textsubscript{2} roten\textsubscript{3} Apfel\textsubscript{4} \textbf{haben}\textsubscript{5} .\textsubscript{6}

    \texttt{(`M', 0, 1, 0, 1)}
    
    \texttt{(`M', 1, 2, 1, 2)}
    
    \textbf{\texttt{(`D', 2, 3, 2, 2)}}
    
    \texttt{(`M', 3, 4, 2, 3)} 
    
    \texttt{(`M', 4, 5, 3, 4)} 
    
    \texttt{(`M', 5, 6, 4, 5)} 
    
    \textbf{\texttt{(`I', 6, 6, 5, 6)} }
    
    \texttt{(`M', 6, 7, 6, 7)}
\end{enumerate}

Relocation of tokens would be a useful label to have for word order errors, which are prevalent in elementary L2 German and Chinese learners. With this label, we could explain why a token is relocated rather than explaining why it is deleted first then explaining why it is inserted.

\section{Guidelines for manual edit extraction Annotation}\label{appendix:manual_extract}

To prepare the data for fine-tuning models to extract atomic edits in German and Mandarin Chinese, we manually annotated $500$ data points for each language. In this section, we discuss the challenges in extracting atomic edits and how we handle them.

The first step is to tokenize sentences. For German, it is straightforward because of white spaces. We use \texttt{SpaCy} for tokenizing German sentences which can single out punctuation marks. For Chinese, sentences cannot be tokenized into words by simply separating characters because many words are not monosyllabic. We choose to use \href{https://github.com/fxsjy/jieba}{Jieba}, which is a fast and accurate Chinese word segmentation module implemented in Python. 

The second step is to use \texttt{SequenceMatcher} from \texttt{difflib} to extract longest edited spans from sentence pairs, which is later used as part of the input for atomic edits. We found that adding rough edits into the input increases the recall of the prompted models. It also accelerates and eases the process of manual annotation.

The third and last step is to get atomic edits. There are four types of edits: replacement, deletion, insertion, and relocation. The challenge lies in how to align words in sentence pairs and extract edits. 

For German, \textbf{replacement} mostly happens between tokens which have similar spelling (e.g., \textit{wolle} and \textit{will}, meaning \textit{want to}) or the same categories (e.g., \textit{zu} and \textit{nach}, meaning \textit{to}). \textbf{Deletion} and \textbf{insertion} can happen to individual tokens or a phrase. When more than one consecutive tokens, for example, X and Y, are deleted or inserted, we determine whether to count them as separate edits or one as a whole depending on whether X and Y form a linguistic constituent (for example, a prepositional phrase \textit{by train}). The edit type \textbf{relocation} is inspired by a common error made by elementary German learners: placing finite verbs or adverbial phrases in the wrong position.\footnote{\label{fn:de.verb.position}German is a verb second language, whose verb second constraint does not hold in embedded clauses. In main clauses, the finite verb occurs in the second position and non-finite verbs occur towards the end of a sentence. In embedded clauses, the finite verb usually appears at the end, after all the non-finite verbs.} To emphasize that the usage of a word is not wrong but its position in a sentence is wrong, tagging such an edit as relocated is more intuitive than tagging it as a deletion followed by an insertion (or an insertion followed by a deletion). 

The introduction of the relocation edit type is not at no cost. It reduces model performance because models tends to predict a relocated token/phrase as deletion plus insertion. It is also challenging because the relocated word should be just placed in a wrong position and have the same dependency in a sentence before and after being relocated. For example, for the sentences in \ref{ex.for}, it is illogical to say that the first sentence is corrected by relocating \textit{for} to the first underline and insert \textit{to} in the original place of \textit{for}. This is because the verb \textit{talk} requires a preposition but the language user mistakenly used \textit{for} instead of \textit{to}. It is not the case that the language user mistakenly put the \textit{for} that should have been before \textit{me} after \textit{talking}. So, it should be the case that \textit{for} is inserted to the position of the blank underline and the \textit{for} after \textit{talking} is replaced by \textit{to}. The correct edits for \ref{ex.for} are given in \ref{ex.for.correct.edits} and the wrong edits are in \ref{ex.for.wrong.edits}

\begin{exe}
    \ex\label{ex.for}
    \textbf{S}: This job is exciting \rule{0.4cm}{0.15mm} me because I like talking \underline{for} different people.\label{ex.source.for}

    \textbf{T}: This job is exciting \underline{for} me because I like talking \underline{to} different people. 

    \ex Good edit extraction\label{ex.for.correct.edits}
    
    \texttt{[`insert', `', `for']}

    \texttt{[`replace', `for', `to']}

    \ex Bad edit extraction\label{ex.for.wrong.edits}
    
    \texttt{[`relocate', `for', `for']}

    \texttt{[`insert', `', `to']}
\end{exe}

On the other hand, the word \textit{essen} in \ref{ex.moechte} is more naturally a relocated token because its relation with the finite modal verb \textit{mo\"ochte} (would like to) and the direct object \textit{vierzig Bananen} (forty bananas) remains unchanged. It is only the position of the word that is changed. 

\begin{enumerate}[label=(\arabic*)]
    \item\label{ex.moechte}
    \begin{enumerate}[label=\alph*.]
    \item Ich m\"ochte \underline{essen} vierzig Bananen.

    \item Ich m\"ochte vierzig Bananen \underline{essen}.

    \texttt{[`relocate', `essen', `essen']}
    \end{enumerate}
\end{enumerate}

For Chinese, deletion and insertion work similarly as in German. Relocation is also useful in Chinese for cases like misplacement of an adverbial phrase or a function word (e.g., \zh{了}).\footnote{\zh{了} is a multi-functional function word and a heteronym. It can express the completion or ongoingness of an action (among its other functions). Its meaning changes based on the position in a sentence it occurs.} However, replacement is not as straightforward in Chinese as in German. For example, verbs in Chinese often come with a resultative complement (e.g., \zh{到}, \zh{完}, or \zh{出}) or other function words to express different states of a verb (e.g., \zh{过}). If only the function word is changed but the verb is not, how should the edit be extracted? We experimented with both ways (with and without verbs) and found that, in either case, \gpt~included the verb when explaining the meaning difference. Hence, for those cases, we always include the unchanged verbs. Similarly, for cases in which a function word is not changed but the verb that the function word is attached to is changed, the edit includes both the verb and the function word (e.g., \texttt{[`replace', `\zh{看成}', `\zh{当成}']}).

\begin{enumerate}[label=(\arabic*)]
    \item\label{ex.guo.wan}
    \begin{enumerate}[label=\alph*.]
    \item \zh{我花了一整天看\underline{过}了这本书。}

    \item \zh{我花了一整天看\underline{完}了这本书。}

    \texttt{[`replace', `\zh{看过}', `\zh{看完}']}
    \end{enumerate}
\end{enumerate}

\noindent Other situations in which we always take longer phrases as edits rather than only the parts being changed are idioms (e.g., \texttt{[`replace', `\zh{心急如坟}', `\zh{心急如焚}']}), formulaic expressions (e.g., \texttt{[`replace', `\zh{总上所述}', `\zh{综上所述}']}), and \textit{de} (\begin{CJK*}{UTF8}{gbsn}\small 的\end{CJK*})$+$ noun as in \begin{CJK*}{UTF8}{gbsn}\small 在这紧急\underline{的情况}下\end{CJK*} (in an emergency situation). 

\section{Prompts for atomic edit extraction}\label{appendix:extraction_prompt}

We use the prompts presented below for atomic edit extraction in German and Chinese. The prompt contains the task instruction followed by possible edit types as well as examples. Special instructions are given to the relocation edit type where the relocated tokens should be the same before and after the edit. In the examples, we demonstrate different edit types and their combinations, showing the models how to deal with a sentence pair with multiple edits. 

\subsection{Extraction prompt for German}\label{appendix:de_extraction_prompt}

\begin{lstlisting}
This is an atomic edit extraction task. Given a pair of German sentences and the edits applied to the first sentence to get the second sentence, your task is to break down the edits to the atomic level (i.e., token level) and assign the edit a label. Be case sensitive. Pay attention to punctuation marks and relocated tokens. Pay attention to phonetic similarity when aligning tokens.

Labels:
1. [replace, original_token, edited_token]
2. [delete, original_token, ""]
3. [insert, "", edited_token]
4. [relocate, original_token, edited_token]: pay attention to tokens that are deleted then added again; the relocated token must be the same before and after the edit.

Examples:
Wie oben schon erwähnt ist die Chance erwisht zurweden zwar gering, aber sie ver handen.
Wie oben schon erwähnt ist die Chance, erwischt zu werden, zwar gering, aber sie ist vorhanden.
Edits:
('replace', 'erwisht zurweden', ', erwischt zu werden ,')
('replace', 'ver handen', 'ist vorhanden')
Atomic edits:
["insert", "", ","]
["replace", "erwisht", "erwischt"]
["replace", "zurweden", "zu werden"]
["insert", "", ","]
["insert", "", "ist"]
["replace", "ver handen", "vorhanden"]

ich haben essen zwei Bananen.
Ich habe zwei Bananen gegessen.
Edits:
('replace', 'ich haben essen', 'Ich habe')
('insert', '', 'gegessen')
Atomic edits:
["replace", "ich", "Ich"]
["replace", "haben", "habe"]
["delete", "essen", ""]
["insert", "", "gegessen"]

Ich habe gegessen zwei Bananen.
Ich habe zwei Bananen gegessen.
Edits:
('delete', 'gegessen', '')
('insert', '', 'gegessen')
Atomic edits:
["relocate", "gegessen", "gegessen"]

Below is the sentence pair for you to work on. Follow the format in the examples strictly. 
{src}
{trg}
Edits:
{edits}
Atomic edits:
\end{lstlisting}

\subsection{Extraction prompt for Chinese}

\begin{CJK*}{UTF8}{gbsn}
You are a Mandarin Chinese teacher. Given a pair of Mandarin Chinese sentences and the edits applied to the input sentence to get the output sentence, your task is to break down the edits to the atomic level (i.e., token level) and assign the edit a label. Pay attention to punctuation marks and relocated tokens.

\noindent Labels:

\noindent 1. [replace, original\_token, editted\_token]

\noindent 2. [delete, original\_token, ""]

\noindent 3. [insert, "", editted\_token]

\noindent 4. [relocate, original\_token1, editted\_token1]: pay attention to tokens that are deleted then added again; the relocated token must be the same before and after the edit.

\noindent Examples:

\noindent 我去市菜场水果买。

\noindent 我去菜市场买水果。

\noindent Edits:

\noindent ("replace", "市菜场水果买", "菜市场买水果")

\noindent Atomic edits:

\noindent ["replace", "市菜场", "菜市场"]

\noindent ["relocate", "水果", "水果"]

\noindent 我吃了早饭今天。

\noindent 我今天吃了早饭。

\noindent Edits:

\noindent ("insert", "今天", "")

\noindent ("delete", "", "今天")

\noindent Atomic edits:

\noindent ["relocate", "今天", "今天"]

\noindent 再子细的学习相关课题后，我意识到了这个问题的严重。

\noindent 在仔细地学习了相关课题后，意识到了这个问题的严重性。

\noindent Edits:

\noindent ("replace", "再子细的", "在仔细地")

\noindent ("insert", "", "了")

\noindent ("insert", "", "我")

\noindent ("insert", "", "性")

\noindent Atomic edits:

\noindent ["replace", "再", "在]

\noindent ["replace", "子细", "仔细"]

\noindent ["replace", "的", "地"]

\noindent ["insert", "", "了"]

\noindent ["insert", "", "我"]

\noindent ["replace", "严重", "严重性"]

\noindent 她打算明儿天的午前去北京。

\noindent 她打算明天上午去北京。

\noindent Edits:

\noindent ("replace", "明儿天的午前", "明天上午")

\noindent Atomic edits:

\noindent ["replace", "明儿天", "明天"]

\noindent ["delete", "的", ""]

\noindent ["replace", "午前", "上午"]

\noindent Below is the sentence pair for you to work on. Follow the format in the examples strictly. 

\noindent \{original\_sentence\}

\noindent \{corrected\_sentence\}

\noindent Edits:

\noindent \{edits\}

\noindent Atomic edits:

\end{CJK*}

\section{Prompts for explanation generation}\label{appendix:expl.gen.prompts}

We use the following prompts for generating grammar error explanations in German and Chinese. 

\subsection{Explanation prompt for German}\label{appendix:de.expl.gen.prompts}

In the prompt for German grammar error explanation, we provide a wide range of error examples, including errors that can only be explained in grammatical terms (e.g., gender/case/number agreement), errors that can be assigned a meaning (e.g., accusative case for time expressions), and errors that are related to collocations (e.g., \textit{am Ende} instead of \textit{im Ende}). 

\begin{lstlisting}
You are given a pair of German sentences and a list of atomic edits. An edit is an error in the first sentence, which is corrected in the second one. Generate a succinct explanation for each error using the template. After each explanation, give the error a type.

Template: The word X is deleted/inserted/replaced by Y/relocated because ...

Example:
Ich habe zwei Bananen für mein Katz gekauft.
Ich habe zwei Bananen für meine Katze gekauft.
Edits:
["replace", "Katz", "Katze"]
["replace", "mein", "meine"]
Explanation:
The word 'Katz' is replaced by 'Katze' because 'Katze' is the correct spelling.
Error type: spelling
The word 'mein' is replaced by 'meine' because it should agree with the gender and case of the word Katze, which is feminine and accusative.
Error type: gender and case

Er fliegt nächster Monat Deutschland.
Er fliegt nächsten Monat nach Deutschland.
Edits:
["insert", "", "nach"]
["replace", "nächster", "nächsten"]
Explanation:
The word 'nach' is inserted because the verb 'fliegen' requires a preposition when expressing a destination and 'nach' is usually used for countries.
Error type: preposition
The word 'nächster' is replaced by 'nächsten' because German uses accusative case for time expressions.
Error type: case

Ich gehe in der Schule.
Ich gehe in die Schule.
Edits:
["replace", "der", "die"]
Explanation:
The word 'der' is replaced by 'die' because the preposition 'in' requires the accusative case of a noun when expressing a direction or destination.
Error type: case

Ich kann heute jogge gehe.
Ich kann heute joggen gehen.
Edits:
["replace", "gehe", "gehen"]
["replace", "jogge", "joggen"]
Explanation:
The word "gehe" is replaced by "gehen" because the verb "kann" requires an infinitive form of the verb "gehen".
Error type: infinitive
The word "jogge" is replaced by "joggen" because the verb "gehen" requires an infinitive form of the verb "joggen".
Error type: infinitive

Ich muss mich zur neuen Umgebung gewöhnen.
Ich muss mich an die neue Umgebung gewöhnen.
Edits:
["replace", "zur", "an"]
["insert", "", "die"]
["replace", "neuen", "neue"]
Explanation:
The word "zur" is replaced by "an" because the verb "gewöhnen" requires the preposition "an".
Error type: preposition
The word "die" is inserted because the noun "Umgebung" requires a determiner and "gewöhnen an" requires accusative case.
Error type: determiner
The word "neuen" is replaced by "neue" because the existence of "die" indicates that the adjective need only weak inflection.
Error type: adjective inflection

Es ist im Ende des Flusses.
Es ist am Ende des Flusses.
Edits:
["replace", "im", "am"]
Explanation:
The word "im" is replaced by "am" because "am" is the correct preposition for the word "Ende".

Below is the sentence pair for you to work on. Focus on the given edit and do not add other atomic edits. Start with the explanation directly.
{src}
{trg}
Edits:
{edit}
Explanation:
\end{lstlisting}

\subsection{Explanation generation prompt for Chinese}\label{appendix:zh.expl.gen.prompts}

In the few-shot prompt for Chinese GEE, we cover the following types of errors, which are commonly seen when we manually annotate the training data for fine-tuning: \textbf{Function word errors}, such as \zh{了}, \zh{们}, \zh{的}/\zh{地}/\zh{得}, and measure words; \textbf{Mis-written words/phrases},\footnote{We call them as mis-written words instead of misspelling because there is no letters or spelling in Chinese writing. Such mistakes can be made by a language user who confuses characters with the same/similar pronunciation, with similar meaning, with similar strokes, or simply remembers the wrong character order in a word.} such as \zh{平果} vs.\ \zh{苹果} and \zh{市菜场} vs.\ \zh{菜市场}; \textbf{Word collocation errors}, such as \zh{做错误} vs.\ \zh{犯错误}; \textbf{Word choice errors}, such as \zh{查找} vs.\ \zh{寻找}.

% Although \citet{rao-etal-2020-overview} does not reveal the levels of the Chinese learners from whom the data were taken, by inspecting the data, we reckon that the data might not be from beginners. 

Mandarin Chinese does not have abundant agreement between words in sentences as German or English. Many errors made by learners are word choice errors. For example, \zh{查找} and \zh{寻找} both have the core meaning of \textit{looking for} but the former emphasizes a systematic and methodological search for data or information while the latter suggests a more intangible search with a sense of exploration. In the example of the word choice error, we show \gpt~that it should explain the meaning of the two words and why one is better than the other in the context. Without such an example, \gpt~returns a generic explanation ``The word X is replace by Y because Y is the correct word to use in the context." which is not helpful for language learners. 

\noindent \textbf{Here begins the prompt:}

\begin{CJK*}{UTF8}{gbsn}

\noindent You are given a pair of Mandarin Chinese sentences and a list atomic edits. An edit is an error in the first sentence, which is corrected in the second one. Generate a succinct explanation for each error using the template. After each explanation, give the error a type.

\noindent Template: The word X is replaced by Y/deleted/inserted/relocated because ...

\noindent Example:

\noindent 昨天我买四只平果们。

\noindent 昨天我买了四个苹果。

\noindent Edits:

\noindent ["insert", "", "了"]

\noindent ["replace", "只", "个"]

\noindent ["replace", "平果", "苹果"]

\noindent ["delete", "们", ""]

\noindent Explanation:

\noindent The word `了' is inserted because `了' indicate the completion of the action `买'.

\noindent Error type: usage of `了'

\noindent The word `只' is replaced with `个' because `个' is the correct measure word for `苹果'.

\noindent Error type: measure word

\noindent The word `平果' is replaced with `苹果' because `苹果' is the correct word for `apple'.

\noindent Error type: miswritten character/word

\noindent The word `们' is deleted because `们' is only used after pronouns or human nouns to indicate plurality.

\noindent Error type: `们'

\noindent 间而说之，他唱地很好。

\noindent 简而言之，他唱得很好。

\noindent Edits:

\noindent ["replace", "间而说之", "简而言之"]

\noindent ["replace", "地", "得"]

\noindent Explanation:

\noindent The word `间而说之' is replaced with `简而言之' because `简而言之' is the correct way of writing the phrase which means `in short' or `in brief'.

\noindent Error type: mis-written character/word

\noindent The word `地' is replaced with `得' because `得' is the correct `de' particle to use when it follows a verb and the word after `得' modifies the verb.

\noindent Error type: "de" particles

\noindent 许多人们做了一差误。

\noindent 许多人犯了一个错误。

\noindent Edits:

\noindent ["replace", "许多人们", "许多人"]

\noindent ["replace", "做", "犯"]

\noindent ["insert", "", "个"]

\noindent ["replace", "差误", "错误"]

\noindent Explanation:

\noindent The word `许多人们' is replaced with `许多人' because when a noun is preceded by a numeral, the plural marker `们' is not needed.

\noindent Error type: `们'

\noindent The word `做' is replaced with `犯' because `犯' is the correct verb to use for the noun `mistake'.

\noindent Error type: verb-object collocation

\noindent The word `个' is inserted because a measure word is needed between the numeral and the noun and `个' is the correct measure word for `错误'.

\noindent Error type: measure word

\noindent The word `差误' is replaced with `错误' because `差误' is not a word in Chinese and `错误' is the correct word for `mistake'.

\noindent Error type: mis-written character/word

\noindent 我在查找我的知音。

\noindent 我在寻找我的知音。

\noindent Edits:

\noindent ["replace", "查找", "寻找"]

\noindent Explanation:

\noindent The word `查找' is replaced with `寻找' because `查找' suggests a systematic and methodological search. It usually means searching for information or data. On the other hand, `寻找' suggests a more intangible search with a sense of exploration. `寻找' fits the context better.

\noindent Error type: word choice

\noindent Below is the sentence pair for you to work on. Focus on the given edit and do not add other atomic edits. Start with the explanation directly.

\noindent \{src\}

\noindent \{trg\}

\noindent Edits:

\noindent \{edit\}

\noindent Explanation:

\end{CJK*}

\section{Data preprocess for German and Chinese}\label{appendix:data.preprocess}

This section describes how the datasets in German and Chinese are preprocessed. 

\subsection{Preprocess German data}\label{appendix:de.data.preprocess}

The Falko dataset \citep{ludeling2008lernerkorpus, reznicek2010falko} contains essays written by German learners whose proficiency levels range from A1 to C1 according to the Common European Framework of Reference for Languages (CEFR).\footnote{\href{https://www.coe.int/en/web/common-european-framework-reference-languages}{The Common European Framework of Reference for Language} (CEFR) is a standard for describing language ability. There are six levels: A1, A2, B1, B2, C1, and C2. C2 is the native speaker level.} The Merlin dataset \citep{boyd-etal-2014-merlin} is a collection of essays written by advanced German speakers from different countries with both native and non-native background. We use Merlin as C2 data. 

Both Falko and Merlin offer two types of grammar error corrections, target hypothesis 1 and target hypothesis 2. Target hypothesis 1 performs minimal correction at the morpho-syntactic level while target hypothesis 2 modifies semantic and pragmatic aspects (e.g., information structure or word choice) of the input text, aiming for a more advanced paraphrase-type correction. For our purpose, we use target hypothesis 1 of each corrected sentence.\footnote{Examples of the target hypothesis 1 and 2 of a corrected sentence can be found in \url{https://gucorpling.org/amir/pdf/Reznicek_et_al.pdf}.}

To prepare the datasets, we first split the paragraphs in Falko and Merlin into sentences by adapting the paragraph alignment algorithm in \citet{thai-etal-2022-exploring} for sentence alignment. We then screened out sentence pairs that: (1) have short sentences (less that 3 tokens); (2) contain ``incomp'' or ``unreadable'' tokens; and (3) have two sentences in the source and one sentence in the target, or vice versa. 

\subsection{Preprocess Chinese data}\label{appendix:zh.data.preprocess}

The data for Chinese GEE is the training split of CGED2017 \citet{rao-etal-2020-overview}. Texts are split into sentences at the end of sentence punctuation (e.g., periods and question marks) and aligned. 

\begin{figure}[]
    \centering
    \includegraphics[width=0.48\textwidth]{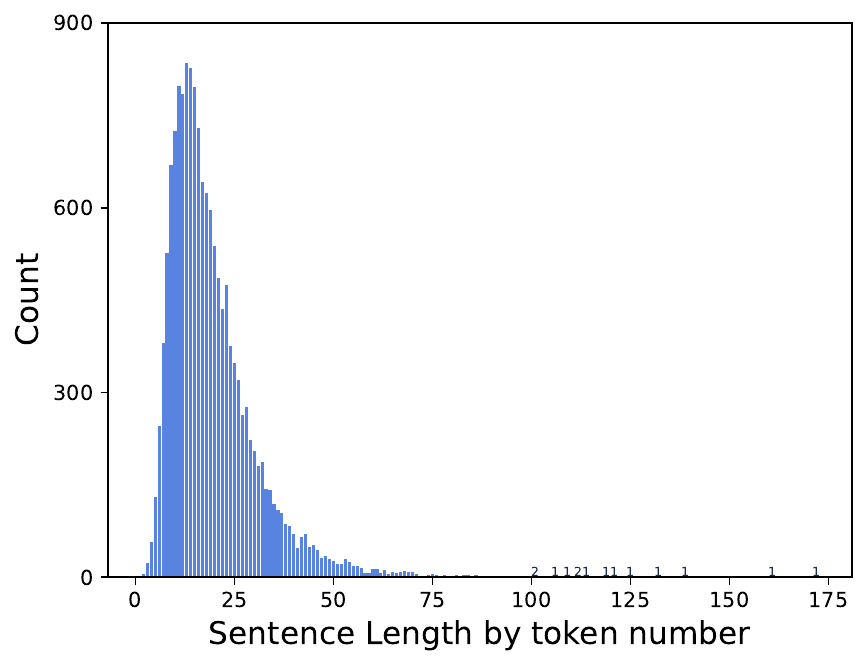}
    \caption{ The sentence length distribution of the data in 2017 CGED training set \citet{rao-etal-2020-overview}. Most of the sentences have less than 50 tokens. For the bars that are invisible in the plot, we add the numbers to them.}
    \label{fig:zh_sent_len_distr}
    \vspace{-0.15in}
\end{figure}

We tokenized the sentence pairs using Jieba and show the length distribution of sentences in \figureref{fig:zh_sent_len_distr}. Clearly, most of the data points have $2$ to $50$ tokens. Each token has on average 1.8 characters. The overly long sentences (over 170 tokens) exist because of the abusive use of commas.\footnote{As a rough reference, \href{https://catalog.ldc.upenn.edu/LDC2016T13}{Chinese Treebank 9.0} \citep{xia2000segmentation} has 132076 sentences and 2084387 tokens, which amounts to roughly 16 tokens per sentence.} For the experiment, we select sentences of length between 5 and 50 tokens. We also remove pairs with the same source and target.

\section{Fine-tune atomic edit extraction models}\label{appendix:fine-tune.detail}

For German we use \llama~and \chatgpt~as the base models and fine-tune them on the 500 training data points in \tableref{tab:dataset.pair.edit.count}. The results show that fine-tuning~\chatgpt~through the OpenAI fine-tuning API with 2 epochs and using temperature $= 0$ at the inference time returns the best performance. It took around 30 mins for fine-tuning. For \llama, we fine-tune the model with QLoRA for 1000 steps using the parameters suggested in \citet{dettmers2023qlora} on one RTX8000. The fine-tuning takes about five hours. Checkpoints are saved every 250 steps. At the inference time, the checkpoint saved at 750 steps with temperature $= 0.01$ performs the best.\footnote{The \texttt{do\_sample} parameter is set to \texttt{False}. The temperature is set to $0.01$ instead of $0$ because the model requires the temperature to strictly be a positive float.} The best performance are reported in \tableref{tab:de_edit_accuracy}.

For Chinese, we fine-tune \llama~and \chatgpt~in the same way as for German. \llama~checkpoints are saved every $100$ steps. It achieves the best performance at $400$ steps. Fine-tuning \chatgpt~for two epochs returns a better performance than one epoch. The best performance of the fine-tuned models are reported in \tableref{tab:zh_edit_accuracy}.

\section{Details on human evaluation}\label{appendix:human.eval}

\begin{figure}[t]
    \centering
    \includegraphics[width=0.48\textwidth]{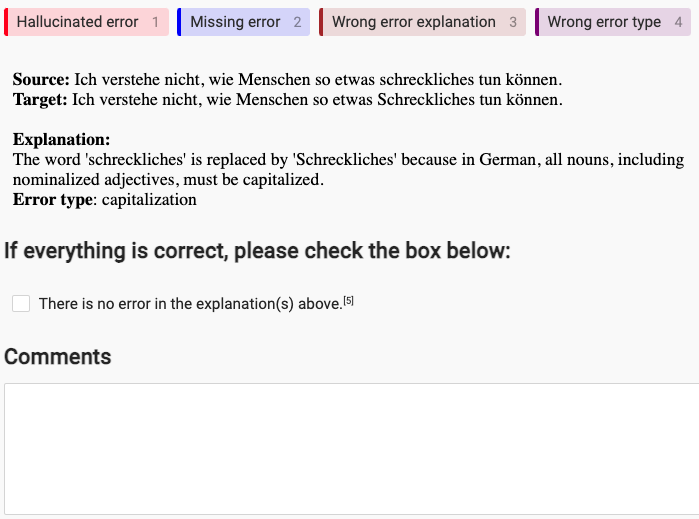}
    \caption{A screenshot of the interface presented to the annotators for explanation evaluation.}
    \label{fig:annotation_interface}
    \vspace{-0.15in}
\end{figure}

We provide further details in addition to the ones discussed in \sectionref{sec:de_expl}. \figureref{fig:annotation_interface} shows the annotation interface for the German and Chinese teachers. The teachers are given detailed instructions for the German \href{https://docs.google.com/presentation/d/18ZT1yXEZ_ttK76cUJHlFY4JfEUiEFBcx97TjRZUTDQ0/edit?usp=sharing}{(link)} and Chinese \href{https://docs.google.com/presentation/d/1cYuvdAyd8xb2mZnlRXsOUSjcl6shX_cqqoYlb9KQZGI/edit?usp=sharing}{(link)} tasks. 

In the annotation task, the teachers are asked to check for four types of mistakes. Concerning \textit{missing error} mistakes, they should be marked either in the source sentence for deleted, replaced, and relocated tokens or in the target sentence for inserted ones. Other mistakes should be marked in the explanations. We asked the annotators not to mark imprecise explanation/error type as wrong but leave a comment on how they can be improved.

\subsection{German annotator agreement}

To evaluate the agreement, we compare the annotations of the 96 pairs and classify them into three categories. \textbf{Fully agree}: if the teachers agree on no mistakes or the same set of mistakes. \textbf{Disagree on missing errors}: if teachers agree on other mistakes but not on \textit{missing errors}. \textbf{Disagree on other mistakes}: if teachers also disagree on mistakes other than \textit{missing errors}. Counts of each category are reported in \figureref{tab:germanagreement}. 

Among the 96 commonly annotated items, the German teachers agree on $81.3\%$ of them for the overall quality (error coverage and explanation quality), and $89.6\%$ of the time, the teachers agree on the quality of the generated edit reasons (sum of the first and second row in \tableref{tab:germanagreement}). 

\begin{table}[]
\centering
\resizebox{\columnwidth}{!}{%
\begin{tabular}{@{}rrr@{}}
\toprule
                             & Count & Percentage \\\midrule
Fully agree                  & 78    & 81.3\%    \\
Disagree on missing errors   & 8     & 8.3\%     \\
Disagree on other mistakes   & 10    & 10.4\%    \\\midrule
Sum                          & 96    & 100\%     \\\bottomrule
\end{tabular}}
\caption{Agreement between two German teachers on 96 sentence pairs. Among the 78 annotated items on which the teachers fully agree with each other, 5 have mistakes and 73 have no mistakes at all.}
\label{tab:germanagreement}
\end{table}

\section{Qualitative analysis of German GEE}\label{appendix:de.expl.quality.analysis}

In this section, we look into the mistakes made by \gpt~and provide detailed analysis of two of them: \textit{wrong error type} and \textit{wrong error explanation}.

\subsection{Mistakes in wrong error type}

Although there are only 12 wrong error type mistakes marked by the German teachers, they present cases where careful design decisions need to be made. We categorize them into six types and discuss two of them here. Examples and their categories are in \tableref{tab:wrong.error.type.cases}.

\noindent \textbf{Case vs.\ Plural} The explanations and error types in the two cases indicate that, given the prompt we used, \gpt~is weak at distinguishing certain nuances in German grammar because it does not leverage the larger context while generating explanations and error types.

In German, the suffix \textit{-n} may occur in two cases (among others): in the plural form of certain nouns or at the end of the dative plural form of a noun if the noun's plural form does not already end in \textit{-n}. In the first case with \textit{Hauspreise}, the language user used \textit{die} as the definite article of \textit{Hauspreise}, which shows that they did not consider the case of the determiner phrase as dative. Moreover, they used \textit{Hauspreise} as part of the subject of the sentence, which further reduces the likelihood that they meant to use \textit{Hauspreise} in its dative case because it is very rare to have a dative determiner phrase as a subject. Hence, the error type should be \textit{plural} or \textit{number}. In the second case with \textit{Menschen}, it is clearly not a \textit{plural} error because \textit{jede/r} takes singular nouns and \textit{bei} only takes dative nouns. The error type should be \textit{case} because the word \textit{Mensch} belongs to the \textit{n}-declination which takes the \textit{-(e)n} suffix in the dative case. Further work should add examples in the prompt for training data to enhance the model ability in distinguishing such nuances.

\noindent \textbf{Misspelling vs.\ Conjugation} While \gpt~judges the errors under this type in \tableref{tab:wrong.error.type.cases} as conjugation errors, our German teachers judged them as misspelling. These three cases beg for an answer to the question: where is the border line between general misspelling due to an oversight and genuinely lack of knowledge of a grammar point (e.g., misspelling vs.\ conjugation)? While we do not have an answer to the question, we suggest that error types should always be the more specific one when an error is on the border line. For a language learner, if an error is made by oversight, they can easily ignore the explanation and error type. If an error is made by lacking of relevant knowledge, they should be reminded by an explanation. Since we do not know why a language learner made such an error, providing the more specific error type is more beneficial. 

\subsection{Mistakes in wrong error explanation}

There are 29 explanations that provide a wrong reason of an error. They can be classified into two groups. The first group has mistakes that can be traced back to a wrongly extracted edit, as shown in the first example in \tableref{tab:wrong.error.expl.cases}. Eleven cases belong to this group. 

The second group has mistakes for miscellaneous reasons. However, there are two reasons that stand out. The first reason is that \gpt~does not consider information from the bigger context when generating explanations. There are 3 such cases and all of them involve a preposition. One example can be found in \tableref{tab:wrong.error.type.cases} under \textit{Case vs.\ Plural}. \tableref{tab:wrong.error.expl.cases} presents another one. In this example, the word \textit{Zimmer} should be in dative not because German needs a dative case to indicate numbers but because the preposition \textit{mit} assigns the noun in the preposition phrase a dative case. The second reason that causes \gpt~to generate four wrong explanations is that it does not have precise knowledge of German verb position. As in the third example in \tableref{tab:wrong.error.expl.cases}, the word \textit{entwickelt} is relocated not because of the reason in the explanation but because a finite verb in an embedded clause should be at the end of it (see Footnote \ref{fn:de.verb.position}). 

\subsection{Overall quality of German GEE}

In the annotation task, the German teachers were told not to mark correct but imprecise explanations/error types as wrong and leave a comment on how they can be improved. In the annotated results, we see only one such comment. That does not mean that the teachers did not leave enough comments because there are abundant comments pointing out errors in the source sentences that are not corrected in the target sentences, comments pointing out that some corrections done in the target sentences are not correct, or comments on how to modify a wrong error explanation. The first author, as a German as second language learner with level C1, has also gone through all the annotated data and found the correct explanations informative and useful. Hence, we can say that the German error explanations generated by \gpt~are judged as fully correct by our German teachers for $93.9\%$ of the time. 

% \yscomment{Other observations: gpt4 follows the instruction very well (format-wise and content-wise). there are 43 sentence pairs in which the errors in the source sentences are not all corrected in the target sentences. but gpt4 ignored those error and focused only on the corrected ones. at the same time, the 43 sentences show that the data quality can be furthre improved.}

% 11032 comment on imprecise 

% Through the error analysis, we saw that the majority of the mistakes in the explanations is caused by the wrong atomic edit extraction (65 in edit description and 11 in edit reason). These mistakes account for $3.82\%$ of the total mistakes. Once the accuracy of the atomic edit extractors is improved, the correct rate of the explanation can be boosted to as high as $97.73\%$. This further proves the neccessity of a good atomic edit extraction model in the task. 

% \begin{table*}[t]
%     \center \footnotesize
%     \setlength{\tabcolsep}{0.5em}
%     \begin{tabular}{p{2.3cm}p{12.8cm}}
%         \toprule
%             Type & Sentence and Explanation with Wrong Error \\
%         \midrule
%             Single-sentence contradiction  &
%         On November 25th, 2023, Glover Teixeira became an American citizen. \newline In November 2020, Teixeira became an American citizen.\\
%         \bottomrule
%     \end{tabular}
%     \caption{todo}\label{tab:ppl-error-analysis}
% \end{table*}

\begin{table*}[]
\fontsize{10}{12}\selectfont
% \small
\centering 
\resizebox{\textwidth}{!}{%
\begin{tabular}{@{}p{20cm}@{}}
\toprule

\multicolumn{1}{c}{\textbf{Overly general}}\\\midrule

Source: Wie es deine Frau und deine Babys? \newline Target: Wie geht es deiner Frau und deinem Baby? \newline The word `' is replaced by `geht' because the correct expression in German is "Wie geht es" to ask how someone is doing. \newline Error type: expression\\[2mm]

Source: Brauchst du etwas hilfe? \newline Target: Brauchst du Hilfe? \newline The phrase `etwas hilfe' is replaced by `Hilfe' because `etwas' is unnecessary when offering help in this context. \newline Error type: redundancy\\[2mm]

Source: Meine Eltren und bin ich in montag fahren in Indien. \newline Target: Meine Eltern und ich fahren am Montag nach Indien. \newline The phrase `in montag' is deleted and replaced by `am Montag' because `am' is the correct preposition for expressing a specific day. \newline Error type: preposition and expression \\\midrule

\multicolumn{1}{c}{\textbf{Misspelling vs.\ Conjugation}}\\\midrule

Source: Ich mochte Farkarte haben. \newline Target: Ich m\"ochte eine Fahrkarte haben. \newline The word `mochte' is replaced by `m\"ochte' because `m\"ochte' is the correct conjugation of the verb ``m\"ogen'' in this context. \newline Error type: conjugation\\[2mm]

Source: Wo m\"ochst du schwimmen gehen? \newline Target: Wo m\"ochtest du schwimmen gehen? \newline The word `m\"ochst' is replaced by `m\"ochtest' because `m\"ochtest' is the correct conjugation of the verb `m\"ochten' in the second person singular. \newline Error type: verb conjugation\\[2mm]

Source: Man ließt es geradezu t\"aglich in der Zeitung, h\"ort es im Radio, oder sieht erschreckende Bilder im Fernsehn. \newline Target: Man liest es geradezu t\"aglich in der Zeitung, hört es im Radio, oder sieht erschreckende Bilder im Fernsehen. \newline The word `ließt' is replaced by `liest' because `liest' is the correct conjugation of the verb `lesen' for the third person singular. \newline Error type: conjugation \\\midrule

\multicolumn{1}{c}{\textbf{Case vs.\ Plural \textit{-(e)n}}}\\\midrule

Source: In den letzten zehn Jahren erhöhten sich die Wohnungs- und Hauspreisen und fast alle m\"ussen einen Kredit aufnehmen, ihre Träume erf\"ollen zu können. \newline Target: In den letzten zehn Jahren erhöhten sich die Wohnungs - und Hauspreise und fast alle müssen einen Kredit aufnehmen, um ihre Träume erf\"ollen zu können. \newline The word `Hauspreisen' is replaced by `Hauspreise' because it should be in the nominative case, which is plural without the `-n' ending \newline Error type: case\\[2mm]
 
Source: Anderseits liegt die Integration jeder Mensch an sie selbst. \newline Target: Anderseits liegt die Integration bei jedem Menschen an ihm selbst. \newline The word 'Mensch' is replaced by `Menschen' because the plural form is needed to agree with the pronoun `jedem'. \newline Error type: plural \\\midrule

\multicolumn{1}{c}{\textbf{Because of wrong atomic edits}}\\\midrule

Source: Obwohl in Stadt X schwerig ist, eine freie Wohnung zu bekommen, gibt es trotzdem viele Möglichkeiten wie man es sie bekommt. \newline Target: Obwohl es in Stadt X schwerig ist, eine freie Wohnung zu bekommen, gibt es trotzdem viele Möglichkeiten wie man sie bekommt. \newline The word `es' is relocated because it should be placed before `in Stadt X' to form the correct sentence structure for expressing a circumstance. \newline Error type: word order\\[2mm]

Source: Bis 30.04 ich muss in neue Wohnung umladen, weil am 01.05 meine neue Arbeit beginn. \newline Target: Bis 30.04. muss ich in die neue Wohnung umladen, weil am 01.05. meine neue Arbeit beginnt.\\The word `ich' is inserted because it is the pronoun that corresponds to the subject performing the action of `müssen'. \newline Error type: subject pronoun \\\midrule

\multicolumn{1}{c}{\textbf{Wrong word category}}\\\midrule

Source: und werde direk zu dir für das Abholen, wäre es dir passen wenn wir den Abends zusammen Essen? \newline Target: Und werde direkt zu dir kommen für das Abholen, würde es dir passen, wenn wir den Abend zusammen essen? \newline The word `wäre' is replaced by `würde' because `würde' is used in conditional sentences to express politeness. \newline Error type: modal verb \\ \midrule

\multicolumn{1}{c}{\textbf{Not understanding German phonology}}\\\midrule

Source: Hast du Seit für mich zum eine Wohnung finden? \newline Target: Hast du Zeit, für mich eine Wohnung zu finden? \newline The word `Seit' is replaced by `Zeit' because `Zeit' is the correct word for time in this context. \newline Error type: homophone \\ \bottomrule

\end{tabular}}
\caption{Twelve cases in which error types are judged as wrong by the German teachers. For each sentence pair, only the explanations with a wrong error type are listed. All grammar errors in the target sentences are from the corpora.}
\label{tab:wrong.error.type.cases}
\end{table*}

\begin{table*}[]
\fontsize{10}{12}\selectfont
% \small
\centering 
\resizebox{\textwidth}{!}{%
\begin{tabular}{@{}p{20cm}@{}}
\toprule

\multicolumn{1}{c}{\textbf{Mistakes that can be traced back to a wrongly extracted edit}}\\\midrule

\textbf{Source}: Dir möchte ich nächten monat nach Amsterdam fahren. 

\textbf{Target}: Ich möchte nächsten Monat nach Amsterdam fahren. \newline 

\textbf{Atomic edits}: \newline 
["replace", "Dir", "Ich"], ["replace", "ich", "n\"achsten"], ["replace", "n\"achten", "Monat"] \newline

\textbf{Explanation}: \newline The word `Dir' is replaced by `Ich' because the subject of the sentence should be `Ich' to express one's own desire to travel. \newline Error type: pronoun \newline The word `ich' is replaced by `nächsten' because the correct word order is `n\"achsten Monat' for `next month'. \newline Error type: word order \newline The word `nächten' is replaced by `Monat' because `n\"achsten' should be followed by the noun `Monat' to form the time expression `n\"achsten Monat'. \newline Error type: word order\\ \midrule

\multicolumn{1}{c}{\textbf{Mistakes due to insufficient consideration of bigger context}}\\\midrule

\textbf{Source}: Ich brauchen ca. 75 m² Wohnung mit 2 Zimmer, Bad, Kuche, toilette und garage.

\textbf{Target}: Ich brauche ca. 75 m² — eine Wohnung mit 2 Zimmern, Bad, Küche, Toilette und Garage. \newline

\textbf{Explanation}:

The word `Zimmer' is replaced by `Zimmern' because in German, the dative case is needed for indicating the number of rooms in this context.

Error type: case \\ \midrule

\multicolumn{1}{c}{\textbf{Mistakes due to imprecise knowledge of verb position in German}}\\\midrule

\textbf{Source}: in dem Sinn, möchte ich dazu sagen, dass integrationsprozes entwickelt sich schneller, wenn man das Gastland nicht als „fremd“ empfindet.

\textbf{Target}: In dem Sinn möchte ich dazu sagen, dass ein Integrationsprozess sich schneller entwickelt, wenn man das Gastland nicht als fremd" empfindet.\newline

\textbf{Explanation}:

The word `entwickelt' is relocated because the verb should be positioned after the subject in a German sentence.

Error type: word order\\\midrule

\multicolumn{1}{c}{\textbf{Mistakes due to other reasons}}\\\midrule

\textbf{Source}: Schauen Sie zuerst ihre Auswahl, sprechen Sie direkt mit dem Verkäufer.

\textbf{Target}: Schauen Sie zuerst ihre Auswahl an, sprechen Sie direkt mit dem Verkäufer. \newline

\textbf{Atomic edit}:

["insert", "", "an"] \newline

\textbf{Explanation}:

The word `an' is inserted because the verb `schauen' requires the preposition `an' when it means 'to look at' something.

Error type: preposition \\

\\\bottomrule

\end{tabular}}
\caption{Examples of wrong error explanations due to different reasons. For each sentence pair, only the explanations with a wrong error explanation are listed. All grammar errors in the target sentences are from the corpora.}
\label{tab:wrong.error.expl.cases}
\end{table*}

\section{Error types generated by \gpt}

\tableref{tab:german.error.types} and \tableref{tab:chinese.error.types} list the frequent error types generated by \gpt~in the German and Chinese GEE task. 

\begin{table*}[]
%\fontsize{10}{12}\selectfont
\centering 
\begin{tabular}{lrrllrr}
\toprule
Error Type & Count & Percent && Error Type & Count & Percent \\
\midrule
punctuation & 520 & 16.48 & & abbreviation & 8 & 0.25 \\
spelling & 470 & 14.89 & & compound noun & 8 & 0.25 \\
capitalization & 353 & 11.19 & & noun form & 7 & 0.22 \\
gender and case & 175 & 5.54 & & extra word & 6 & 0.19 \\
preposition & 163 & 5.16 & & syntax & 6 & 0.19 \\
word order & 157 & 4.97 & & adjective & 6 & 0.19 \\
case & 119 & 3.77 & & adverb & 6 & 0.19 \\
determiner & 100 & 3.17 & & word form & 6 & 0.19 \\
adjective inflection & 71 & 2.25 & & verb tense & 6 & 0.19 \\
verb conjugation & 62 & 1.96 & & noun & 5 & 0.16 \\
conjunction & 59 & 1.87 & & spelling and capitalization & 5 & 0.16 \\
pronoun & 39 & 1.24 & & tense & 5 & 0.16 \\
conjugation & 33 & 1.05 & & comparative & 5 & 0.16 \\
verb form & 30 & 0.95 & & formatting & 5 & 0.16 \\
word choice & 30 & 0.95 & & word formation & 5 & 0.16 \\
redundancy & 30 & 0.95 & & possessive pronoun & 4 & 0.13 \\
plural & 29 & 0.92 & & preposition and case & 4 & 0.13 \\
infinitive & 29 & 0.92 & & time expression & 4 & 0.13 \\
unnecessary word & 26 & 0.82 & & possessive & 4 & 0.13 \\
vocabulary & 26 & 0.82 & & auxiliary verb & 4 & 0.13 \\
subject-verb agreement & 25 & 0.79 & & demonstrative pronoun & 4 & 0.13 \\
article & 22 & 0.70 & & idiomatic expression & 4 & 0.13 \\
verb & 20 & 0.63 & & missing subject & 4 & 0.13 \\
adjective agreement & 20 & 0.63 & & past participle & 4 & 0.13 \\
reflexive pronoun & 19 & 0.60 & & spacing & 4 & 0.13 \\
gender & 16 & 0.51 & & separable verb & 4 & 0.13 \\
expression & 13 & 0.41 & & negation & 4 & 0.13 \\
subject & 13 & 0.41 & & modal verb & 4 & 0.13 \\
compound word & 12 & 0.38 & & terminology & 4 & 0.13 \\
missing word & 11 & 0.35 & & relative pronoun & 4 & 0.13 \\
adjective form & 11 & 0.35 & & singular/plural & 4 & 0.13 \\
plural form & 11 & 0.35 & & gender agreement & 4 & 0.13 \\
subject omission & 10 & 0.32 & & compound verb & 4 & 0.13 \\
verb choice & 10 & 0.32 & & verb agreement & 4 & 0.13 \\
missing verb & 8 & 0.25 & & spelling and inflection & 4 & 0.13 \\
translation & 8 & 0.25 & & compound separation & 4 & 0.13 \\
\bottomrule
\end{tabular}
\caption{A distribution over error types in German grammatical error explanations (3156 total points, types with 4 or more datapoints considered). Overall, we observe a wide variety of error types.}
\label{tab:german.error.types}
\end{table*}

\begin{table*}[]
%\fontsize{10}{12}\selectfont
\centering 
\begin{tabular}{lrrllrr}
\toprule
Error Type & Count & Percent && Error Type & Count & Percent \\
\midrule
word choice & 588 & 39.65 & & extraneous word & 7 & 0.47 \\
redundancy & 120 & 8.09 & & unnecessary `\zh{的}' & 7 & 0.47 \\
word order & 101 & 6.81 & & preposition usage & 7 & 0.47 \\
missing word & 55 & 3.71 & & subject omission & 6 & 0.40 \\
miswritten character/word & 52 & 3.51 & & `\zh{们}' & 5 & 0.34 \\
usage of `\zh{了}' & 44 & 2.97 & & missing particle & 5 & 0.34 \\
"de" particles & 31 & 2.09 & & redundant character & 5 & 0.34 \\
preposition & 24 & 1.62 & & redundant `\zh{的}' & 5 & 0.34 \\
redundant word & 22 & 1.48 & & emphasis & 5 & 0.34 \\
conjunction & 21 & 1.42 & & particle usage & 4 & 0.27 \\
omission & 20 & 1.35 & & redundant phrase & 4 & 0.27 \\
verb-object collocation & 19 & 1.28 & & auxiliary verb & 4 & 0.27 \\
word omission & 18 & 1.21 & & modal verb & 4 & 0.27 \\
unnecessary word & 17 & 1.15 & & missing verb & 4 & 0.27 \\
sentence structure & 15 & 1.01 & & unnecessary particle & 4 & 0.27 \\
usage of `\zh{的}' & 14 & 0.94 & & conjunction/connective & 3 & 0.20 \\
extra word & 11 & 0.74 & & missing words & 3 & 0.20 \\
grammar & 9 & 0.61 & & idiomatic expression & 3 & 0.20 \\
missing information & 9 & 0.61 & & aspect particle & 3 & 0.20 \\
conjunction usage & 8 & 0.54 & & unnecessary character & 3 & 0.20 \\
missing subject & 8 & 0.54 & & adverb usage & 3 & 0.20 \\
measure word & 8 & 0.54 & & expression & 3 & 0.20 \\
negation & 8 & 0.54 & & unnecessary use of `\zh{的}' & 3 & 0.20 \\
\bottomrule
\end{tabular}
\caption{A distribution over error types in Chinese grammatical error explanations (1483 total points, types with 3 or more datapoints considered). Overall, we observe a wide variety of error types.}
\label{tab:chinese.error.types}
\end{table*}

\end{document}